\documentclass[lettersize,journal]{IEEEtran}

\usepackage[colorlinks,urlcolor=blue,linkcolor=blue,citecolor=blue]{hyperref}
\usepackage{color,array}

\usepackage{graphicx}

\usepackage{graphicx}
\usepackage{pifont}
\usepackage{amsmath}
\usepackage{amssymb}
\usepackage{mathrsfs}

\usepackage{algpseudocode}  
\usepackage{makecell}
\usepackage{float}  
\usepackage{subfigure}  
\usepackage[normalem]{ulem}
\useunder{\uline}{\ul}{}
\usepackage{multirow}
\usepackage{booktabs}
\usepackage{subcaption}
\usepackage{algorithm}
\usepackage{algpseudocode}
\usepackage{xcolor}
\usepackage{algcompatible}
\usepackage[justification=centering]{caption}
\begin{document}

\title{Pattern-Matching Dynamic Memory Network for Dual-Mode Traffic Prediction} 

\author{Wenchao Weng, Mei Wu, Hanyu Jiang, Wanzeng Kong, \IEEEmembership{Senior Member, IEEE},\\ Xiangjie Kong, \IEEEmembership{Senior Member, IEEE}, and Feng Xia, \IEEEmembership{Senior Member, IEEE}
\thanks{This work was supported in part by the "Pioneer" and "Leading Goose" R\&D Program of Zhejiang under Grant 2024C01214, and in part by the National Natural Science Foundation of China under Grant 62072409. (corresponding author: Xiangjie Kong.)}
\thanks{Wenchao Weng and Xiangjie Kong are with the College of Computer Science and Technology, Zhejiang Universityof Technology, Hangzhou 310023, China (e-mail: 111124120010@zjut.edu.cn; xjkong@ieee.org).}
\thanks{Mei Wu and Hanyu Jiang are with the Hangzhou Dianzi University ITMO Joint Institute, Hangzhou Dianzi University, Hangzhou 310018, China (e-mail: 222320007@hdu.edu.cn; 22320324@hdu.edu.cn).}
\thanks{Wanzeng Kong is with College of Computer Science, Hangzhou Dianzi University, Hangzhou 310018, China (e-mail: kongwanzeng@hdu.edu.cn).
}
\thanks{Feng Xia is with School of Computing Technologies, RMIT University, Melbourne, VIC 3000, Australia (e-mail: f.xia@ieee.org).
}
}


\maketitle

\begin{abstract}
In recent years, deep learning has increasingly gained attention in the field of traffic prediction. Existing traffic prediction models often rely on GCNs or attention mechanisms with \(O(N^2)\) complexity to dynamically extract traffic node features, which lack efficiency and are not lightweight. Additionally, these models typically only utilize historical data for prediction, without considering the impact of the target information on the prediction.
To address these issues, we propose a Pattern-Matching Dynamic Memory Network (PM-DMNet). PM-DMNet employs a novel dynamic memory network to capture traffic pattern features with only \(O(N)\) complexity, significantly reducing computational overhead while achieving excellent performance. The PM-DMNet also introduces two prediction methods: Recursive Multi-step Prediction (RMP) and Parallel Multi-step Prediction (PMP), which leverage the time features of the prediction targets to assist in the forecasting process. Furthermore, a transfer attention mechanism is integrated into PMP, transforming historical data features to better align with the predicted target states, thereby capturing trend changes more accurately and reducing errors.
Extensive experiments demonstrate the superiority of the proposed model over existing benchmarks. 
The source codes are available at: \url{https://github.com/wengwenchao123/PM-DMNet}.
\end{abstract}



\begin{figure}[ht]

\end{figure}

\section{Introduction}

\IEEEPARstart{W}{ith} the development of society and technology, there has been a significant increase in vehicles within cities, as well as the growing popularity of services like shared bicycles and ride-hailing platforms such as Uber and Didi. This expansion has broadened the application of urban traffic management by governments and heightened public transportation demands. However, issues such as limited resources and inadequate scheduling systems have increasingly highlighted challenges in traffic management and the imbalance of transportation demand.
As a result, accurate traffic forecasting has become a crucial issue in fields such as traffic management, urban planning, and the sharing economy. Precise traffic prediction enables governments to better allocate social resources to maintain urban transportation operations. It also allows companies to distribute resources such as shared bicycles and taxis to areas with high demand, thereby avoiding their idle presence in low-demand areas. This approach can reduce energy consumption and passenger waiting times.

In recent years, researchers have conducted extensive studies in traffic prediction to promote the development of intelligent transportation systems.
Early traffic prediction methods utilized statistical approaches for prediction. Auto-regressive (AR), Moving Average (MA), and Auto-Regressive Integrated Moving Average (ARIMA) models \cite{box1970distribution}, as the most representative classical statistical methods, have been extensively employed in traffic prediction. Additionally, machine learning techniques represented by Support Vector Regression (SVR) \cite{wu2004travel} and Kalman filters \cite{guo2014adaptive} have also been applied to traffic prediction to achieve more accurate predictions and handle more complex sequences. However, these methods require data to exhibit stationarity to be effective, which limits their ability to capture the intricate non-linear spatio-temporal correlations present in traffic condition.

  
In recent years, the advancements of deep learning in domains such as Computer Vision and Natural Language Processing have motivated researchers to explore its application in traffic prediction for improved outcomes. Early deep learning prediction models conceptualized urban traffic as images and segmented them into grids. Convolutional Neural Networks (CNNs) \cite{zhang2017deep} were employed to analyze spatial correlations within these grids, while Recurrent Neural Networks (RNNs) \cite{bai2020adaptive,li2023dynamic,li2018diffusion} or CNNs \cite{wu2019graph,yu2018spatio} were utilized to capture temporal dependencies. However, the structure of the transportation network can be viewed as a topological graph, containing non-Euclidean attributes. CNNs only extract features from the surrounding nodes and cannot capture features from other locations across space. As Graph Convolutional Networks (GCNs) \cite{kipf2016semi} are effective in handling non-Euclidean structures, it has been widely applied in the field of transportation \cite{weng2023decomposition, wu2019graph, li2023dynamic}. Additionally, attention mechanisms \cite{pdformer,zheng2020gman,guo2023self} have been incorporated for spatio-temporal feature modeling.

\begin{figure}[t]
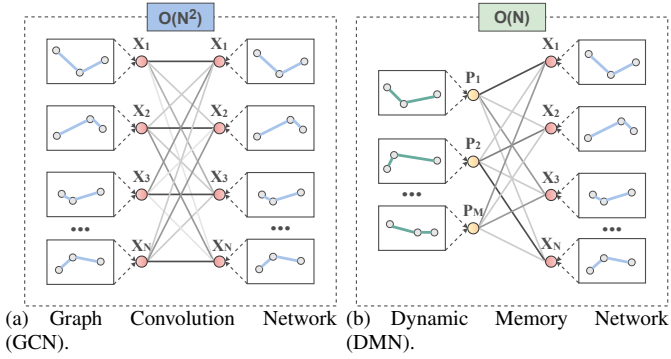

	\centering  
	\subfigbottomskip=2pt 
	\subfigcapskip=-5pt 
	\subfigure[{Graph Convolution Network (GCN).}]{
        \label{Graph Convolution Network (GCN).}
		\includegraphics[width=0.47\linewidth]{GCN.pdf}}
	\subfigure[{Dynamic Memory Network (DMN).}]{
        \label{Dynamic Memory Network (DMN).}
		\includegraphics[width=0.47\linewidth]{PM.pdf}}
	\caption{Comparison between GCN and DMN. As $M$ is constant, the time complexity of GCN and DMN is $O(N^2)$ and $O(N)$, respectively.}
   \label{Comparison between GCN and DMN. As $M$ is constant, the time complexity of GCN and DMN is $O(N^2)$ and $O(N)$, respectively.}
\end{figure}

\begin{figure}[t]
	\centering  
	\subfigbottomskip=2pt 
	\subfigcapskip=-5pt 
	\subfigure[{Nodes with different traffic patterns.}]{
        \label{Nodes with different traffic patterns.}
		\includegraphics[width=0.9\linewidth]{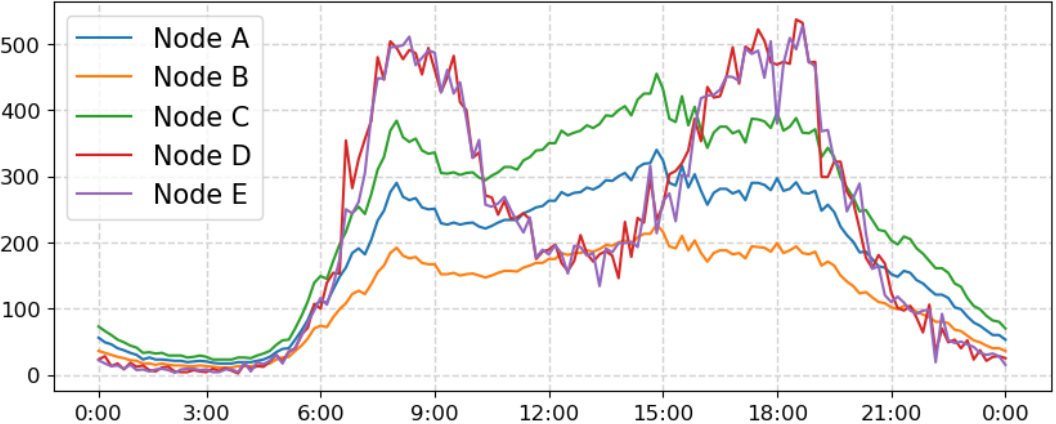}}
	\subfigure[{Similar historical traffic conditions, different future traffic conditions.}]{
        \label{Similar historical traffic conditions, different future traffic conditions.}
		\includegraphics[width=0.9\linewidth]{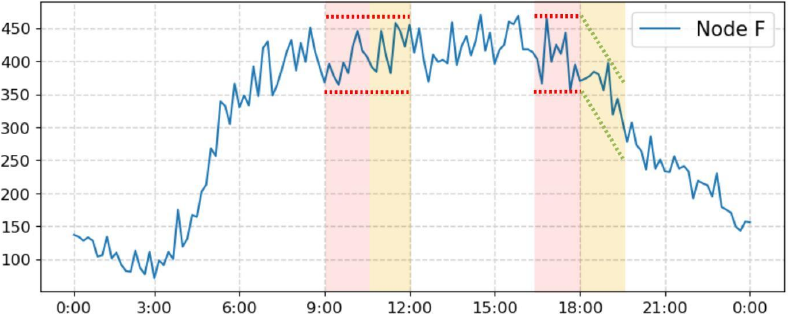}}
	\caption{The findings about traffic data.}
   \label{The Findings about traffic data.}
\end{figure}

However, current methods still possess the following limitations: 

1) Lack of Effective Traffic Feature Extraction: Traffic data inherently exhibits complex spatio-temporal correlations. To capture these spatio-temporal correlations, researchers have employed GCN to capture spatial relationships between nodes, achieving significant success.
As shown in Figure \ref{Graph Convolution Network (GCN).}, current methods require evaluating the correlations between all pairs of nodes to dynamically generate the graph structure and then use GCN to extract spatio-temporal correlations \cite{li2023dynamic,weng2023decomposition}, resulting in an $O(N^2)$ computational complexity. 
However, in practical scenarios, the structure of transportation networks often exhibits sparsity, meaning that nodes are only correlated with a subset of other nodes, and most nodes do not have correlations with each other. As illustrated in Figure \ref{Nodes with different traffic patterns.}, Nodes A, B, and C exhibit evident correlations, representing a specific traffic pattern, while Nodes D and E signify another traffic pattern. Computing similarities between Nodes A, B, C, and Nodes D, E would be meaningless and resource-intensive. 
Recent studies \cite{kong2022exploring,kong2024exploring,fang2023spatio} have focused on reducing computational complexity, but they each come with limitations. For instance, STWave \cite{fang2023spatio} introduces an MS-ESGAT (Multi-Scale Edge-based Spatial Graph Attention) mechanism to achieve linear complexity. However, this method relies heavily on predefined graph structures, making it unsuitable for scenarios where no predefined graph is available. 

2) Uncertainty in Predicting Trend Changes: Figure \ref{Similar historical traffic conditions, different future traffic conditions.} illustrates two sets of historical data and their corresponding prediction targets, where the red segment represents historical data and the yellow segment represents the prediction target. As shown, the left side's historical data and corresponding prediction targets remain within a stable trend channel. However, on the right side, although the historical data is also within a stable trend channel, the corresponding prediction target shifts into a downward trend channel. 
This indicates that relying solely on historical data for prediction makes it challenging to capture such trend shifts. Although current studies \cite{zhang2023multi,bai2020adaptive,ye2022learning} have proposed various methods to extract spatiotemporal features, they rely exclusively on historical data to model traffic conditions, leading to limitations in accurately capturing the trend changes of prediction targets.


To address the above issues, a novel Pattern-Matching Dynamic Memory Network (PM-DMNet) model for traffic prediction is proposed in this paper. For the first challenge, a Dynamic Memory Network (DMN) is designed to extract pattern features from nodes. Specifically, a learnable memory matrix is defined to learn representative traffic patterns within the traffic conditions. The traffic features input to the model are then used in conjunction with these embeddings to compute a pattern attention matrix, which facilitates the extraction of features from the most similar traffic patterns. Simultaneously, the DMN dynamically adjusts the representative traffic patterns at each time point by combining time embeddings with memory embeddings, thus avoiding issues related to traffic pattern homogenization.
Moreover, as illustrated in Figure \ref{Comparison between GCN and DMN. As $M$ is constant, the time complexity of GCN and DMN is $O(N^2)$ and $O(N)$, respectively.}, compared to the high computational complexity of GCN, which is $O(N^2)$, this method reduces the computational complexity to $O(N)$, significantly enhancing computational efficiency.

To address the second challenge, two prediction methods are designed: Recurrent Multi-step Prediction (RMP) and Parallel Multi-step Prediction (PMP). RMP uses the traditional recursive approach, where predictions are made during the decoding phase by recursively utilizing the time features and extracted hidden features for the target time points.  PMP directly uses the time features for the target time points and the hidden features extracted from historical data for prediction.  To mitigate the errors caused by discrepancies between historical data and prediction targets, a novel Transition Attention Mechanism is introduced in PMP. Specifically, this attention mechanism leverages the inherent periodicity in traffic data by integrating the input data, its time features, and the time features of the prediction targets. This transforms the hidden states to better align with the conditions of the target time points. This method enhances the adaptability of the extracted latent features to the prediction target states, improving accuracy. Furthermore, PMP reduces the required computation time compared to RMP, as it does not involve recursion, and it also enhances prediction performance.

In summary, the contributions of this paper can be summarized as follows:

\begin{itemize}
\item[$\bullet$] We present a new traffic prediction model, named Pattern Matching Dynamic Memory Network (PM-DMNet). This model can achieve both Parallel Multi-step Prediction (PMP) and Recurrent Multi-step Prediction (RMP) in the decoder stage depending on the requirements. Compared to RMP, PMP avoids the cyclic recursion process, thereby enhancing computational efficiency.


\item[$\bullet$] We propose a novel Dynamic Memory Network (DMN) module designed to learn inherent representative traffic patterns within the data associated with each node. By employing a pattern matching approach, this module identifies and extracts traffic pattern features most similar to the input data while effectively reducing computational overhead.

\item[$\bullet$] We introduce a new Transfer Attention Mechanism (TAM). TAM transforms the existing historical hidden states into latent states aligned with the prediction target features, mitigating the error caused by the discrepancy between historical data and prediction targets.

\item[$\bullet$] Experimental results on ten authentic datasets substantiate that our proposed framework significantly outperforms state-of-the-art methods across all datasets.





\end{itemize}

\begin{figure*}[!ht]
  \centering
  \includegraphics[width=\linewidth]{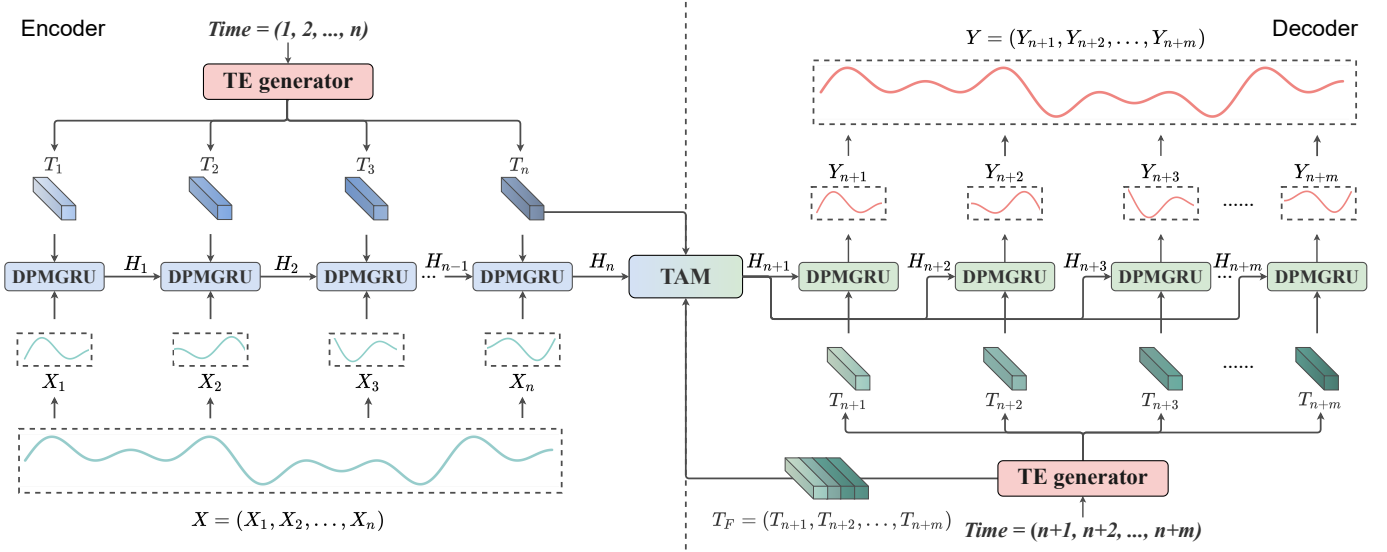}
  \caption{Overview of PM-DMNet structure.}
  \label{Overview of PM-DMNet structure.}
\end{figure*}

\section{Related work}

\subsection{Spatio-Temporal Prediction}
As one of the most representative tasks in spatio-temporal prediction, researchers employed a myriad of methodologies to model the spatio-temporal characteristics within traffic condition. STGCN \cite{song2020spatial} leveraged GCN and predefined matrices to capture spatial correlations between nodes, employing Gate CNNs to model such spatial dependencies. DCRNN \cite{li2018diffusion} integrated diffusion convolution with GRU to model the spatio-temporal relationships inherent in traffic condition. MTGNN \cite{wu2020connecting} utilized adaptive embeddings to generate an adaptive graph structure, capturing spatial correlations among diverse nodes. CCRNN \cite{ye2021coupled} introduced a novel graph convolutional structure termed as CGC and employed a hierarchical coupling mechanism, linking upper-layer graph structures with underlying ones to extract temporal-spatial features. 
GMAN \cite{zheng2020gman} harnessed three distinct attention mechanisms to capture the spatio-temporal characteristics present in traffic condition.
MPGCN \cite{kong2022exploring} utilized GCN to identify mobility patterns at bus stops through clustering and employed GCN2Flow to predict passenger flow based on various mobility patterns. Building on the foundation of MPGCN, MPGNNFormer \cite{kong2024exploring} designed a STGNNFormer to extract both temporal and spatial dependencies. 
Although these spatiotemporal prediction models have achieved notable success, the GCNs and attention mechanisms they use often require \( O(N^2) \) or even higher complexity, resulting in substantial computational costs.

\subsection{Neural Memory Network}

The Memory Network \cite{weston2015memory} introduced an external memory mechanism, enabling it to better handle and utilize long-term information. Memory networks have found extensive applications in the domains of natural language processing and machine translation. MemN2N \cite{sukhbaatar2015end} introduced a novel end-to-end memory network framework that facilitates its straightforward application in real-world environments. Kaiser et al. \cite{kaiser2016learning} proposed memory networks with the capability to adapt to various zero-shot scenarios. Mem2seq \cite{madotto2018mem2seq} integrated multi-hop attention mechanisms with memory networks, enabling their deployment in dialog systems. MemAE \cite{gong2019memorizing} explored the application of memory networks in video anomaly detection tasks, subsequent studies \cite{lv2021learning} validating the feasibility of this approach. MTNet \cite{chang2018memory} endeavored to apply memory networks in multi-variate time series prediction, yielding promising results. In the most recent advancements, PM-MemNet \cite{lee2022learning} devised a novel Graph Convolutional Memory Network (GCMem) to model the spatio-temporal correlations inherent in given traffic condition. Additionally, MegaCRN \cite{jiang2023spatio}, inspired by memory network principles, designed a Meta-Graph Learner to construct dynamic graphs, addressing temporal-spatial heterogeneities. Although memory networks have been applied in traffic prediction, they still require integration with other feature extraction methods (e.g., GCN) to perform effectively.

Unlike previous spatio-temporal prediction models, PM-DMNet uses a dynamic memory network to extract traffic pattern features, achieving superior performance while reducing complexity to \( O(N) \), which significantly lowers computational costs. Additionally, prior research overlooks the impact of time features corresponding to the prediction targets on the targets themselves. PM-DMNet fully considers this characteristic and designs two prediction methods to utilize these time features, leading to successful outcomes.

\section{PRELIMINARIES}
\subsection{Temporal Indexing Function}

\begin{table}[ht]
\caption{Example of time index transformation}
\centering
\captionsetup{justification=centering}
\begin{tabular}{ccc}
\toprule[1pt]%
Time           & d(t)    & w(t)     \\ \toprule[1pt]%
Monday,00:05   & 0:05:00 & Monday   \\
Monday,01:00   & 1:00:00    & Monday   \\
Thursday,01:00 & 1:00:00    & Thursday \\ \toprule[1pt]%
\end{tabular}
\label{Example of time index transformation}
\end{table}

Given that traffic condition is collected at regular time intervals, each set of traffic condition possesses unique and systematic temporal information. To harness these temporal characteristics effectively, we employ a temporal indexing function to extract time-related information. Let $d(t)$ and $w(t)$ represent the intra-daily and weekly indexing functions, respectively. These functions transform the temporal information of the traffic condition into corresponding intra-daily and weekly time-related attributes. For specific examples, refer to Table \ref{Example of time index transformation}.

\subsection{Traffic Prediction}

The objective of traffic prediction is to utilize historical traffic condition to forecast future traffic condition.

We represent the traffic condition $X_{t}\in \mathbb{R}^{N \times C}$ for $N$ nodes in the road network at time $t$, where $C$ is the dimensionality of traffic condition, signifying $C$ types of traffic condition. We model the historical traffic condition $X=[X_{1},X_{2},...,X_{n}]\in \mathbb{R}^{n\times N\times C}$ over the past $n$ time steps using the model $f$ to predict the traffic condition $Y=[Y_{n+1},Y_{n+2},...,Y_{n+m}]\in \mathbb{R}^{m\times N\times C}$ for the future $m$ time steps, which can be expressed as:
\begin{equation}
[X_{1},X_{2},...,X_{n}]\xrightarrow{f}[Y_{n+1},Y_{n+2},...,Y_{n+m}]
\end{equation}
In addition, The corresponding actual values are represented by $\hat{Y}=[\hat{Y}_{n+1},\hat{Y}_{n+2},...,\hat{Y}_{n+m}]\in \mathbb{R}^{m\times N\times C}$,

\begin{figure}[ht]
  \centering
  \includegraphics[width=0.8\linewidth]{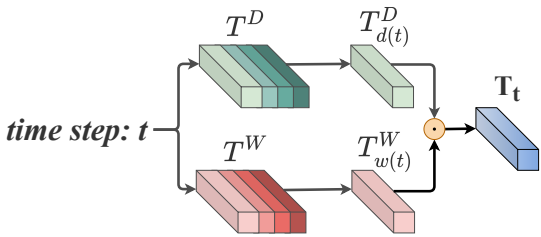}
  \caption{The construction of TE generator.}
  \label{The construction of TE generator.}
\end{figure}

\section{Model Archithecture}

Figure \ref{Overview of PM-DMNet structure.} illustrates the comprehensive architecture of PM-DMNet, which comprises a Time Embedding Generator (TE Generator), Dynamic Pattern Matching Gated Recurrent Unit (DPMGRU), and Transfer Attention Mechanism (TAM).  In the subsequent sections, we will provide a detailed exposition of each module.

\subsection{Time Embedding Generator}

Traffic condition is influenced by people's travel habits and lifestyles, exhibiting clear temporal  such as rush hours during mornings and evenings. To fully leverage temporal features, we introduce two independent embedding pools $T^{D}\in \mathbb{R}^{N_{d}\times p}, T^{W}\in \mathbb{R}^{N_{w}\times p}$ to learn features for intra-daily and weekly patterns. Here,  $N_{d}$ represents the number of time slots in a day, and $N_{w}=7$ represents the number of days in a week.	As depicted in Figure \ref{The construction of TE generator.}, based on the time information $t$, we derive the intra-daily index $d(t)$ and the weekly index $ w(t)$. Utilizing $d(t)$ and $w(t)$, we obtain the intra-daily time feature embedding $T^{D}_{d(t)}$ and the weekly time feature embedding $T^{W}_{w(t)}$ corresponding to the specific time point. Ultimately, these $T^{D}_{d(t)}\in \mathbb{R}^{p}$ and $T^{W}_{w(t)}\in \mathbb{R}^{p}$ are integrated to yield a combined time embedding, which can be expressed as follows:
\begin{equation}
T_{t}=T^{D}_{d(t)} \odot T^{W}_{w(t)}
\end{equation}
where $\odot$ denotes the hadamard product.

\subsection{Dynamic Memory Network}

The memory module incorporates a learnable memory matrix $P=[P^{1},P^{2},...,P^{M}]\in \mathbb{R}^{M \times p}$, where symbolizes a unique traffic pattern. To dynamically adjust the memory matrix, thereby avoiding pattern singularization and adapting to the prevailing traffic conditions at time $t$, we integrate the current time embedding $T_{t}$ with $P$. This fusion can be represented as:
\begin{equation}
P_{t} =P \odot T_{t}
\end{equation}
where $P_{t} \in \mathbb{R}^{M \times p}$ represents the memory network module at time $t$. Through training, $P_{t}$ can learn the most representative traffic patterns at time $t$. By integrating the time embedding $T_{t}$ dynamically, the model can adjust its memory $P_{t}$ to better capture evolving traffic patterns and conditions over time.

\begin{figure}[ht]
  \centering
  \includegraphics[width=\linewidth]{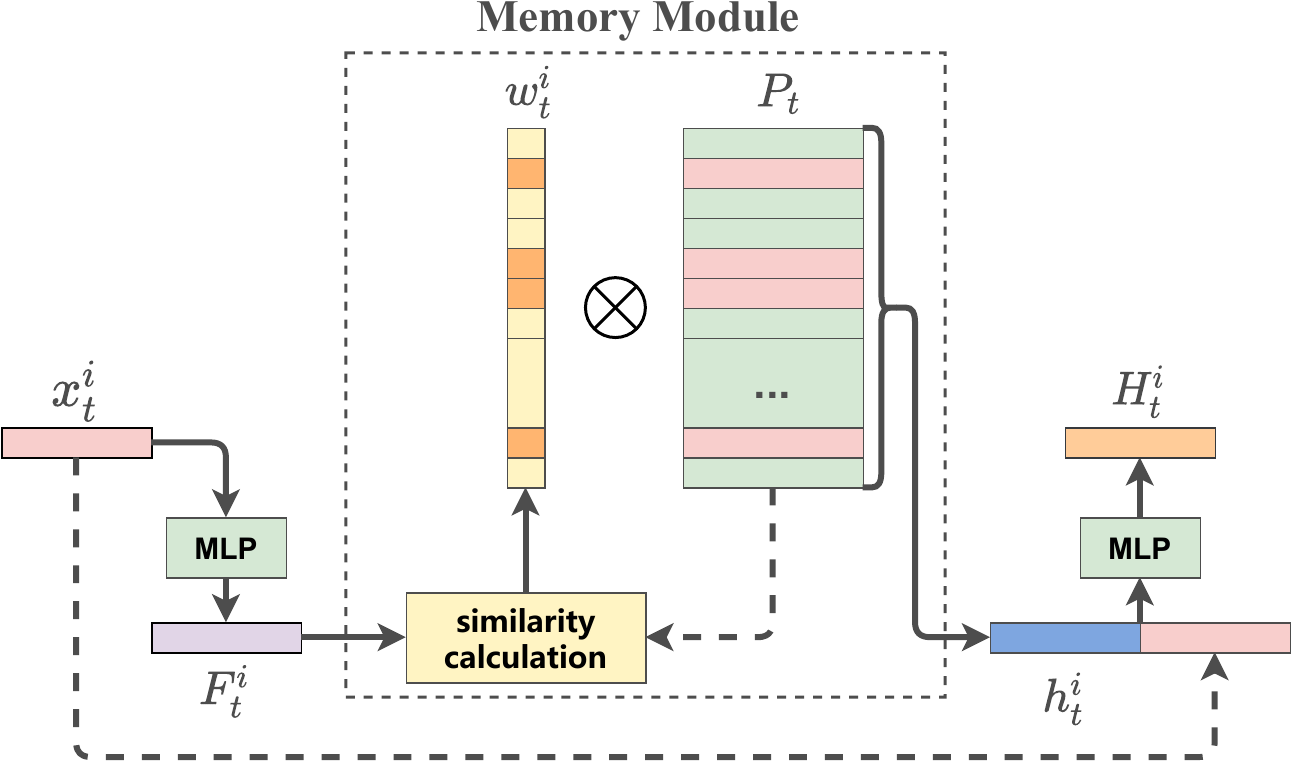}
  \caption{Dynamic memory network.}
  \label{Dynamic Memory Network.}
\end{figure}

As shown in Figure \ref{Dynamic Memory Network.}, we extract dynamic signals from the traffic condition, which can be represented as:
\begin{equation}
F^{i}_{t} = MLP(x^{i}_{t})
\end{equation}
where $F^{i}_{t}\in \mathbb{R}^{p}$ represents the dynamic signal extracted from the traffic condition $x^{i}_{t}$ at node $i$. It is used to query the memory matrix for the traffic pattern most similar to $x^{i}_{t}$.

Afterwards, the similarity weight between $ F^{i}_{t} $ and the memory matrix $ P_{t} $ is computed through a similarity calculation:
\begin{equation}
w^{i}_{t} = softmax(F^{i}_{t}P^{T}_{t})
\end{equation}
where $w^{i}_{t}\in \mathbb{R}^{M}$ represents the similarity weight vector.

Subsequently, $P_{t}$ is linearly transformed to obtain the pattern features corresponding to various traffic patterns. It is then multiplied with the similarity weight vector $w^{i}_{t}$ to extract the pattern features most similar to $x^{i}_{t}$, as follows:
\begin{equation}
h^{i}_{t}=w^{i}_{t}P_{t}
\end{equation}
where $h_{t}\in \mathbb{R}^{M\times F_{out}}$ represents the pattern features in the memory matrix $P_{t}$, and $h^{i}_{t}$ represents the extracted traffic pattern features.

Finally, the residual connection is employed to concatenate $h^{i}_{t}$ and $x^{i}_{t}$ for extracting hidden features:
\begin{equation}
H^{i}_{t}=(h^{i}_{t}||x^{i}_{t})\Theta
\end{equation}
where $\Theta \in \mathbb{R}^{F_{in} \times F_{out}} $ represents learnable parameters. All node hidden states $H^{i}_{t}$ are aggregated into $H_{t}=(H^{1}_{t},H^{2}_{t},...,H^{N}_{t})$, serving as the final output of the dynamic memory network.

\subsection{Node Adaptive Parameter Learning}

To enable each node to learn its unique traffic pattern, enhancing the model's robustness and effectiveness, we utilize two parameter matrices to optimize the learnable parameters $\Theta$. Specifically, we use the node embedding matrix  $E \in \mathbb{R}^{N \times d}$ and the weight pool  $W \in \mathbb{R}^{d \times F_{in} \times F_{out}}$ to generate  $\Theta \in \mathbb{R}^{N \times F_{in} \times F_{out}}$, which can be expressed as:
\begin{equation}
\begin{aligned}
  H^{i}_{t}&=(h^{i}_{t}||x^{i}_{t})\Theta\\
           &=(h^{i}_{t}||x^{i}_{t})E\cdot W
\end{aligned}
\end{equation}
where $\cdot$ represents the multiplication of matrices in different dimensions. From the perspective of an individual node, $E$ provides $d$ independent traffic patterns, and the node adjusts $W$ in a data-driven way to assign appropriate weights to each pattern. These weights are combined to create the node’s unique traffic pattern.

\subsection{Dynamic Pattern Matching Gated Recurrent Unit}

To capture the spatio-temporal features inherent in traffic condition, we integrate the gated recurrent unit (GRU) with a dynamic memory network to construct a framework that encapsulates both temporal dynamics and spatial correlations.   Specifically, we replace the MLP layer in the GRU with a dynamic memory network, resulting in the Dynamic Pattern Matching Gated Recurrent Unit (DPMGRU). Mathematically, DPMGRU can be formulated as:
\begin{equation}
\begin{aligned}
  r_{t}&=\sigma(\vartheta_{r*G}(x_{t}||H_{t-1}))\\
  u_{t}&=\sigma(\vartheta_{u*G}(x_{t}||H_{t-1}))\\
  h_{t}&=tanh(\vartheta_{h*G}(x_{t}||u_{t}\odot H_{t-1}))\\
  H_{t}&=r_{t}\odot H_{t-1} +(1-r_{t})\odot h_{t}
\end{aligned}
\end{equation}
where $ X_t$ and $ H_t$ denote the input and output at time step $ t$, respectively. $\sigma$ represents the $sigmoid$ activation function. $r$ and $ u$ correspond to the reset gate and update gate, respectively. $*G$ denotes the dynamic memory network module, while $\vartheta_{r}, \vartheta_{u}, \vartheta_{h}$ are the learnable parameters associated with the relevant memory network module.

\subsection{Transfer Attention Mechanism}

To mitigate the discrepancy between historical data and the prediction target leading to errors, we employ a transfer attention mechanism to transform the learned hidden features from historical data. Specifically, we first linearly transform the encoder's output $H_{n} \in \mathbb{R}^{N \times D}$, historical time embedding $T_{n} \in \mathbb{R}^{p}$, and future embeddings  $T_{F}=(T_{n+1},T_{n+2},...,T_{n+m}) \in \mathbb{R}^{m \times p}$ into queries, keys, and values, represented as:
\begin{equation}
Q=\forall(H_{n},T_{F})W^{Q},K=\forall(H_{n},T_{n})W^{K},V=\forall(H_{n},T_{n})W^{V}
\end{equation}
where $W^{Q},W^{K},W^{V}\in \mathbb{R}^{(D+p) \times d_{k}}$ serve as learnable parameters, and $\forall()$ denotes a broadcasting operation. Subsequently, the transfer attention can be expressed as:

\begin{equation}
\begin{aligned}
H_{TA}&=attention(H,T_{F},T_{n})\\
     &=softmax(\frac{QK^{T}}{\sqrt{d_{K}}}V)
\end{aligned}
\end{equation}

Finally, the feature fusion between $H_{n}$ and $H_{TA} \in \mathbb{R}^{m \times N \times D}$ is achieved using residual connections to obtain the input for the decoder:
\begin{equation}
H_{out}= MLP(\forall(H_{n},H_{TA}))
\end{equation}
where $H_{out}=(H_{n+1},H_{n+2},...,H_{n+m}) \in \mathbb{R}^{m \times N \times D}$ correspond to the hidden features from time points $n+1$ to $n+m$ for the prediction target. By employing $T_{N}$ and $T_{F}$, these features undergo transfer learning to adapt more effectively to the state of the prediction target time points.

\subsection{Encoder-Decoder Architecture}

\begin{figure}[ht]
  \centering
  \includegraphics[width=\linewidth]{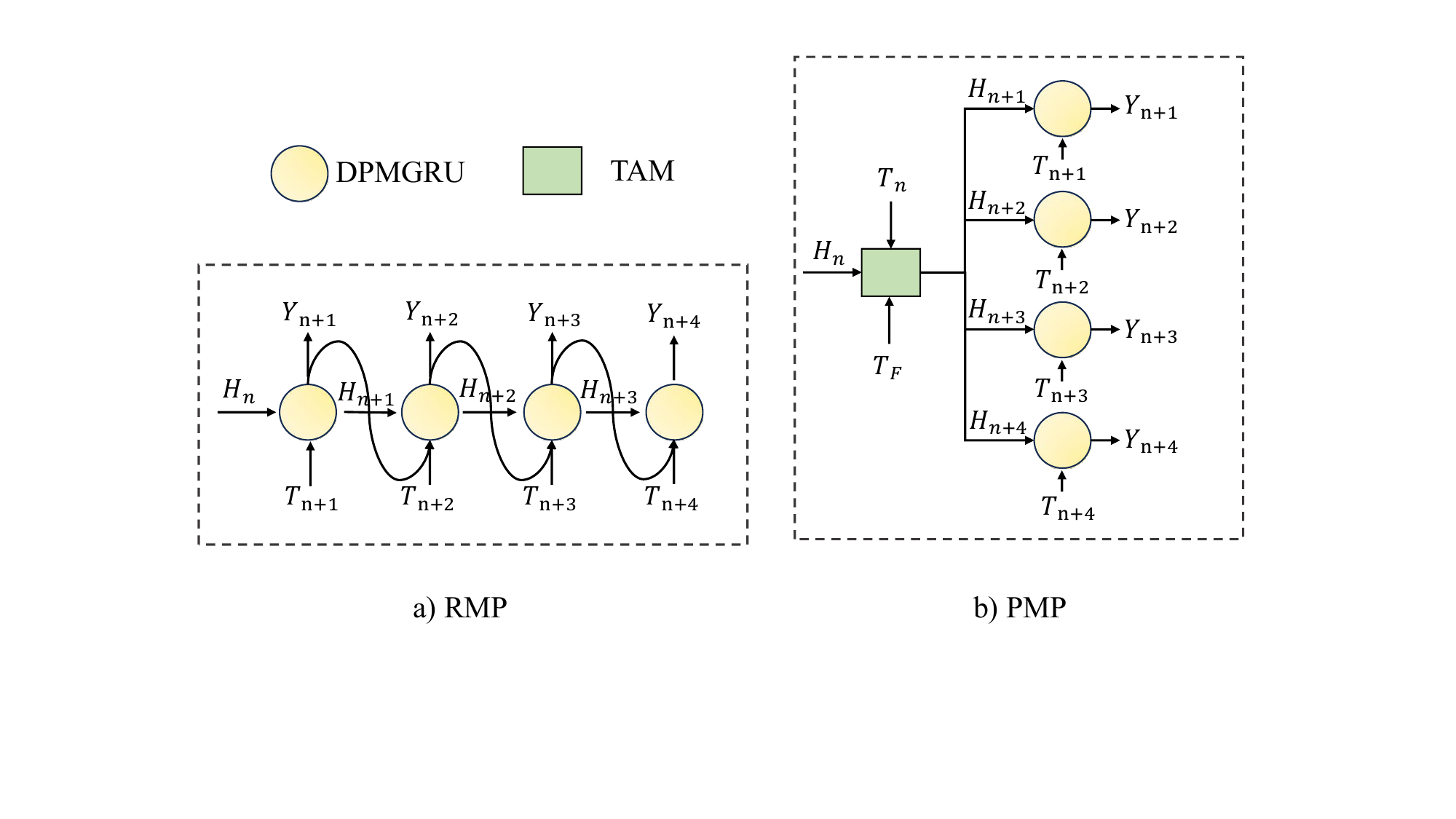}
  \caption{Comparison of Recurrent Multi-step Prediction (RMP) and Parallel Multi-step Prediction (PMP).}
  \label{Comparison of Recurrent Multi-step Prediction (RMP) and Parallel Multi-step Prediction (PMP).}
\end{figure}




The traditional encoder-decoder architecture typically employs the Recurrent Multi-step Prediction (RMP) method for forecasting. However, recurrent decoding has inherent limitations, including: (i) error accumulation due to recurrent predictions, and (ii) the sequential nature of recursion, which restricts the model’s ability for parallel computation, thus limiting the improvement of inference speed. \cite{lin2023segrnn} demonstrates that Parallel Multi-step Prediction (PMP) methods can achieve comparable or even better results than RMP when appropriate techniques are applied. Therefore, two variants are designed to implement and investigate these prediction methods:

\textbf{PM-DMNet(R)} As illustrated in Figure \ref{Comparison of Recurrent Multi-step Prediction (RMP) and Parallel Multi-step Prediction (PMP).}(a), PM-DMNet(R) employs the classic Recurrent Multi-step Prediction (RMP) method, where $Y_{t}$ is derived through single-step prediction. Subsequently, the predicted $Y_{t}$ serves as the input for predicting $Y_{t+1}$, iterating this process until the complete prediction output is obtained.

\textbf{PM-DMNet(P)} Inspired by \cite{lin2023segrnn}, PM-DMNet(P) adopts the Parallel Multi-step Prediction (PMP) method. As shown in Figures \ref{Comparison of Recurrent Multi-step Prediction (RMP) and Parallel Multi-step Prediction (PMP).}(b), during the decoding phase, the encoder’s output $H_n$ is first processed through TAM to obtain $H_{out}$, which alleviates the discrepancy between historical data and prediction targets, aligning it more closely with the state of the prediction targets. Subsequently, $H_{out} = (H_{n+1}, H_{n+2}, \ldots, H_{n+m})$ and $T_{F} = (T_{n+1}, T_{n+2}, \ldots, T_{n+m})$ are segmented and input into PDMGRU to predict the corresponding targets $Y = (Y_{n+1}, Y_{n+2}, \ldots, Y_{n+m})$. Since recursive compilation is not required, the prediction targets $Y$ can be predicted in parallel, avoiding the issue of accumulating prediction errors with recursion steps. 

The details of model training and prediction are presented in Algorithm \ref{alg:Training algorithm}.

\begin{algorithm}[htb]
  \caption{ Training algorithm of PM-DMNet.}
  \label{alg:Training algorithm}
  \renewcommand{\algorithmicrequire}{\textbf{Input:}}
  \renewcommand{\algorithmicensure}{\textbf{Output:}}
  \begin{algorithmic}[1]
    \REQUIRE
      The traffic dataset $O$, 
      encoder’s function $f_{en}(\cdot)$,
      decoder’s function $f_{de}(\cdot)$,
      TAM's function $f_{tam}(\cdot)$,
      prediction type $T$,
      scheduled sampling function $f_{ss}(\cdot)$

    \REPEAT
    \STATE select a input $X\in R^{n\times N\times C}$, label $\hat{Y}\in R^{m \times N\times C}$, time information $t$, initialize hidden state $H_0$.
    
      
        \STATE compute $T_{t}=T^{D}_{d(t)} \odot T^{W}_{w(t)}$
        \FOR{$i$ in $1,2,...,n$ }
            \STATE compute $H_{i} = f_{en}(X[i,...],H_{i-1},T_{i})$
        \ENDFOR
        
        \IF {$T$= PM-DMNet(P)} 
            \STATE Initialize a zero tensor $Y_{in} \in R^{m\times N \times C}$ as the input to the decoder.
            \STATE compute $H_{out} = f_{tam}(H_{n},T_{N},T_{F})$
            \STATE compute $Y = f_{de}(Y_{in},H_{out},T_{F})$
        \ENDIF  
        
        \IF {$T$= PM-DMNet(R)} 
        \STATE set iter = 1;
        \STATE Initialize a zero tensor $Y_{in} \in R^{N \times C}$ as the input to the decoder.
        \FOR{$q$ in $1,2,...,m$ }
            \STATE compute $Y[q,:] = f_{de}(Y_{in},H_{m+q-1},T_{n+q})$
            \STATE compute $\varepsilon_{i} = f_{ss}(iter)$
            \STATE generate a random number $\mu \sim \mathbf{N}(0, 1)$.
            \IF { $\mu$ $<$ $\varepsilon_{i}$} 
            \STATE $Y_{in}= \hat{Y}[q,...]$.
            \ELSE
            \STATE $Y_{in}= Y[q,...]$
            \ENDIF  
        \ENDFOR
        \ENDIF

      \STATE Calculate loss $L$ by using MAE.
      \STATE Update model parameters according to loss $L$.
      \UNTIL{convergence of the model is achieved}

        
    \ENSURE
       learned model.
  \end{algorithmic}
\end{algorithm}

\section{Experimental Setup}
\subsection{Datasets \& Settings}

In this section, experiments are conducted on ten real-world datasets to validate the effectiveness of the proposed PM-DMNet. The datasets used are categorized into four types: bike demand datasets include  NYC-Bike14 \cite{zhang2017deep}, NYC-Bike15 \cite{yao2019revisiting}, and NYC-Bike16 \cite{ye2021coupled}; taxi demand datasets include  NYC-Taxi15 \cite{yao2019revisiting} and NYC-Taxi16 \cite{ye2021coupled}; traffic flow datasets include PEMSD4 \cite{guo2019attention}, PEMSD7 \cite{song2020spatial}, and PEMSD8 \cite{guo2019attention}; and traffic speed datasets include PEMSD7(M) and PEMSD7(L) \cite{yu2018spatio}. Detailed information about the datasets and the training set divisions can be found in Table \ref{Statistics of datasets.}. Moreover, Unlike traffic flow and traffic speed datasets, the traffic demand datasets have two dimensions: 'Pick-up' and 'Drop-off'. We set $n = 12$ historical time steps to predict $m = 12$ future time steps.

\begin{table}[ht]
\captionsetup{justification=centering}
\caption{Statistics of datasets.}
\label{Statistics of datasets.}
\resizebox{\linewidth}{!}{
\begin{tabular}{ccccccc}
\toprule[1pt]%
Data type                      & Datasets  & Nodes & Time steps & Time Range        & Time interval & Train/Val/Test \\ \toprule[1pt]%
\multirow{3}{*}{Bike Demand}   & NYC-Bike14  & 128   & 4392       & 04/2014 - 09/2014 & 1 hour        & 7/1/2          \\
                               & NYC-Bike15  & 200   & 2880       & 01/2015 - 03/2015 & 30 min        & 7/1/2          \\ 
                               & NYC-Bike16  & 250   & 4368       & 04/2016 - 06/2016 & 30 min        & 7/1.5/1.5      \\
                               \toprule[1pt]%
\multirow{2}{*}{Taxi Demand}   & NYC-Taxi15   & 200   & 2880       & 01/2015 - 03/2015 & 30 min        & 7/1/2          \\ 
                               & NYC-Taxi16  & 266   & 4368       & 04/2016 - 06/2016 & 30 min        & 7/1.5/1.5      \\
                                \toprule[1pt]%
\multirow{3}{*}{Traffic Flow}  & PEMSD4    & 307   & 16992      & 01/2018 - 02/2018 & 5min          & 6/2/2          \\
                               & PEMSD7    & 883   & 28224      & 05/2017 - 08/2017 & 5min          & 6/2/2          \\
                               & PEMSD8    & 170   & 17856      & 07/2016 - 08/2016 & 5min          & 6/2/2          \\ \toprule[1pt]%
\multirow{2}{*}{Traffic Speed} & PEMSD7(M) & 228   & 12672      & 05/2012 - 06/2012 & 5min          & 6/2/2          \\
                               & PEMSD7(L) & 1026  & 12672      & 05/2012 - 06/2012 & 5min          & 6/2/2          \\ \toprule[1pt]%
\end{tabular}
}
\end{table}

All experiments are conducted on a server equipped with an NVIDIA GeForce GTX 4090 GPU. The Adam optimizer is used for model optimization, and the Mean Absolute Error (MAE) is adopted as the loss function. The hyper-parameter settings for the model under the two different prediction methods, such as the temporal embedding dimension $p$, node embedding dimension $d$, memory network matrix dimension $M$, batch size, and learning rate , are detailed in Table \ref{Model Hype-rparameter Settings.}. During training, an early stopping strategy is employed to terminate training and prevent over-fitting. Additionally, a scheduled sampling strategy \cite{bengio2015scheduled} is applied to PM-DMNet(R) to enhance its robustness. 

\begin{table}[ht]
\captionsetup{justification=centering}
\caption{Model hyper-parameter settings.}
\label{Model Hype-rparameter Settings.}
\resizebox{\linewidth}{!}{
\begin{tabular}{cccccccc}
\toprule[1pt]
\multirow{2}{*}{Datasets} & \multicolumn{2}{c}{PM-DMNet(P)} & \multicolumn{2}{c}{PM-DMNet(R)} & \multirow{2}{*}{$M$} & \multirow{2}{*}{batchsize} & \multirow{2}{*}{learning rate} \\ \cline{2-5}
                         & $p$              & $d$              & $p$              & $d$              &                    &                            &                     \\ \toprule[1pt]
NYC-Bike14                 & 20             & 10             & 20             & 10             & 10                 & 64                         & 0.03                \\
NYC-Bike15                 & 12             & 6              & 12             & 6              & 10                 & 64                         & 0.03                \\ 
NYC-Bike16                 & 20             & 10             & 20             & 10             & 10                 & 64                         & 0.03                \\ \toprule[1pt]
NYC-Taxi15                  & 20             & 10             & 20             & 10             & 10                 & 64                         & 0.03                \\ 
NYC-Taxi16                 & 20             & 10             & 20             & 10             & 10                 & 64                         & 0.03                \\ \toprule[1pt]
PEMSD4                   & 24             & 12             & 20             & 10             & 10                 & 64                         & 0.03                \\
PEMSD7                   & 24             & 12             & 24             & 12             & 10                 & 64                         & 0.03                \\
PEMSD8                   & 20             & 10             & 12             & 6              & 10                 & 64                         & 0.03                \\ \toprule[1pt]
PEMSD7(M)                & 8              & 4              & 10             & 5              & 10                 & 64                         & 0.03                \\
PEMSD7(L)                & 16             & 8              & 20             & 10             & 10                 & 64                         & 0.03                \\ \toprule[1pt]
\end{tabular}
}
\end{table}


\subsection{Baselines}

To compare performance, the following 24 baselines with official code are compared with PM-DMNet:

1) Traditional Models:
\begin{itemize}
\item[$\bullet$] HA \cite{hamilton2020time}: It utilizes historical averages to iteratively predict the future.

\item[$\bullet$]ARIMA \cite{box1970distribution}: It integrates moving averages into an auto-regressive model.

\item[$\bullet$] VAR \cite{williams2003modeling}: It is a statistical model capable of capturing spatial dependencies.

\end{itemize}

2) Machine Learning Models:
\begin{itemize}
\item[$\bullet$] SVR \cite{wu2004travel}: It uses support vector machines for prediction.

\item[$\bullet$] XGBoost \cite{chen2016xgboost}: It is a classical and widely adopted machine learning model.
\end{itemize}

3) Deep Learning Models:
\begin{itemize}
\item[$\bullet$] LSTM \cite{sutskever2014sequence}: It makes predictions through iterations.

\item[$\bullet$] TCN \cite{bai2018empirical}: It employs causal convolutions and dilated convolutions to capture temporal correlations.

\item[$\bullet$] STGCN \cite{zhang2017deep}: It uses graph convolution and one-dimensional convolutional neural networks to separately extract spatial and temporal correlations.

\item[$\bullet$] STGCN \cite{yu2018spatio}: It combines TCN with GCN to extract spatio-temporal dependencies.

\item[$\bullet$] DCRNN \cite{li2018diffusion}: It combines diffusion convolution and GRU to extract spatiotemporal correlations.

\item[$\bullet$] STG2Seq \cite{bai2019stg2seq}: It captures temporal dependencies from both long-term and short-term perspectives.

\item[$\bullet$] GWN \cite{wu2019graph}: It integrates gated TCN and adaptive graph GCN to capture spatiotemporal dependencies.

\item[$\bullet$] ASTGCN \cite{guo2019attention}: It performs attention mechanism analysis on spatio-temporal convolutions to extract dynamic spatio-temporal correlations.

\item[$\bullet$] LSGCN \cite{huang2020lsgcn}: It uses graph convolutional networks and a novel cosine graph attention network to capture long-term and short-term spatial dependencies.

\item[$\bullet$] STFGNN \cite{song2020spatial}: It designs a spatio-temporal fusion graph to capture local spatio-temporal correlations.

\item[$\bullet$] STSGCN \cite{song2020spatial}: It constructs a three-dimensional graph for graph convolution to capture spatio-temporal correlations between nodes.

\item[$\bullet$] MTGNN \cite{wu2020connecting}: It employs self-learned adjacency matrices and a time convolution module to capture spatio-temporal correlations between different variables.

\item[$\bullet$] CCRNN \cite{bai2020adaptive}: It designs a Coupled Layer-wise Graph Convolution for prediction.

\item[$\bullet$] STFGNN \cite{song2020spatial}: It designs a spatio-temporal fusion graph to capture local spatio-temporal correlations.

\item[$\bullet$] STGODE \cite{fang2021spatial}: It leverages neural ODE to reconstruct GCN, alleviating the over-smoothing problem in deep GCNs.

\item[$\bullet$] GTS \cite{shang2021discrete}: It learns the graph structure among multiple time series and simultaneously makes predictions using GNN.

\item[$\bullet$] ESG \cite{ye2022learning}: It designs an evolving structure learner to construct a series of adjacency matrices. These matrices not only receive information from the current input but also maintain the hidden states of historical graph structures.

\item[$\bullet$] MVFN \cite{zhang2023multi}: It uses graph convolution and attention mechanisms to extract local and global spatial features. Additionally, it employs multi-channel and separable temporal convolutional networks to extract overall temporal features.

\item[$\bullet$] STWave \cite{fang2023spatio}: It uses the DWT algorithm to decouple traffic data for modeling. Additionally, it designs a novel local graph attention network to efficiently and effectively model dynamic spatial correlations.

\item[$\bullet$] MegaCRN \cite{jiang2023spatio}: It designs a Meta-Graph Learner to construct dynamic graphs, addressing temporal-spatial heterogeneities.

\end{itemize}

\begin{table*}[ht]
\caption{Performance comparison between PM-DMNet and the baselines on five traffic demand datasets. The best results are highlighted in bold, and the second-best results are underlined.}
\captionsetup{justification=centering}
\label{Performance comparison between PM-DMNet and the baselines on five traffic demand datasets.}
 \resizebox{\linewidth}{!}{
\begin{tabular}{cccccccccccccccc}
\toprule[1pt]%
\multirow{2}{*}{Method} & \multicolumn{3}{c}{NYC-Bike16}                        & \multicolumn{3}{c}{NYC-Taxi16}                        & \multicolumn{3}{c}{NYC-Bike14}                        & \multicolumn{3}{c}{NYC-Bike15}                        & \multicolumn{3}{c}{NYC-Taxi15}                          \\ \cline{2-16} 
                        & RMSE            & MAE             & CORR            & RMSE            & MAE             & CORR            & RMSE            & MAE             & CORR            & RMSE            & MAE             & CORR            & RMSE             & MAE             & CORR            \\ \toprule[1pt]%
XGBoost                 & 4.0494          & 2.4689          & 0.4107          & 21.1994         & 11.6806         & 0.4416          & 10.3137         & 4.8228          & 0.3322          & 8.1780          & 2.7175          & 0.1289          & 44.1421          & 14.8994         & 0.2195          \\
DCRNN                   & 3.2274          & 1.8973          & 0.6601          & 14.8318         & 8.4835          & 0.6671          & 6.3259          & 2.7483          & 0.5184          & 3.8320          & 1.2645          & 0.2844          & 16.6155          & 5.6424          & 0.4909          \\
STGCN                   & 3.7829          & 2.2076          & 0.5933          & 14.6473         & 7.8435          & 0.7257          & 8.5412          & 3.5833          & 0.4481          & 5.6169          & 1.6101          & 0.2529          & 28.1391          & 9.1844          & 0.3454          \\
STG2Seq                 & 3.7843          & 2.2055          & 0.5413          & 19.2077         & 10.4925         & 0.5389          & 10.8561         & 4.4999          & 0.3751          & 8.2462          & 2.3272          & 0.1855          & 39.4318          & 12.8251         & 0.3764          \\
STSGCN                  & 2.8846          & 1.7538          & 0.7126          & 10.9692         & 5.8299          & 0.8242          & 7.8272          & 3.2998          & 0.4656          & 5.4722          & 1.6086          & 0.2373          & 28.0221          & 8.9541          & 0.3695          \\
MTGNN                   & 2.7791          & 1.6595          & 0.7353          & 10.9472         & 5.9192          & 0.8249          & 6.3548          & 2.8172          & 0.5154          & 3.9407          & 1.2947          & 0.2640          & 18.1113          & 5.9255          & 0.5284          \\
CCRNN                   & 2.7674          & 1.7133          & 0.7333          & 9.8744          & 5.6636          & 0.8416          & 7.4890          & 3.5197          & 0.4861          & 4.4359          & 1.5249          & 0.2681          & 23.0052          & 8.5411          & 0.4049          \\
GTS                     & 2.9258          & 1.7798          & 0.6985          & 12.7511         & 7.2095          & 0.7348          & 6.7053          & 2.9446          & 0.5044          & 4.1698          & 1.3632          & 0.2654          & 17.8672          & 6.0408          & 0.4462          \\
ESG                     & 2.6727          & 1.6129          & 0.7449          & 8.9759          & 5.0344          & 0.8592          & 6.3503          & 2.7972          & 0.5175          & 3.8054          & 1.2293          & 0.2756          & 16.7635          & 5.5279          & 0.5247          \\
MVFN                    & 2.6981          & 1.6565          & 0.7380          & 8.7953          & 4.9433          & 0.5607          & 6.4116          & 2.8228          & 0.5131          & 3.9282          & 1.2928          & 0.2793          & 16.2687          & 5.5613          & 0.5296          \\
MegaCRN                 & 2.7480          & 1.6321          & 0.7425          & 8.7082          & 4.9082          & 0.8619          & 6.3258          & 2.8005          & 0.5185          & 3.9459          & 1.2681          & 0.2836          & 15.4985          & 5.2107          & 0.5398          \\ \toprule[1pt]%
PM-DMNet(P)             & \textbf{2.5631} & \textbf{1.5566} & \textbf{0.7709} & {\ul 8.4699}    & {\ul 4.7682}    & {\ul 0.8674}    & {\ul 5.8790}    & {\ul 2.5687}    & \textbf{0.5274} & \textbf{3.5302} & \textbf{1.1678} & \textbf{0.2849} & \textbf{14.6360} & \textbf{4.8126} & \textbf{0.5509} \\
PM-DMNet(R)             & {\ul 2.5964}    & {\ul 1.5667}    & {\ul 0.7638}    & \textbf{8.4659} & \textbf{4.7635} & \textbf{0.8675} & \textbf{5.8656} & \textbf{2.5582} & {\ul 0.5246}    & {\ul 3.7118}    & {\ul 1.1947}    & {\ul 0.2700}    & {\ul 14.7843}    & {\ul 4.8629}    & {\ul 0.5429}    \\ \toprule[1pt]%
\end{tabular}
}
\end{table*}
 \begin{figure*}[ht]
	\centering  
	\subfigbottomskip=2pt 
	\subfigcapskip=-5pt 
	\subfigure[{RMSE on NYC-Bike16}]{
		\includegraphics[width=0.185\linewidth]{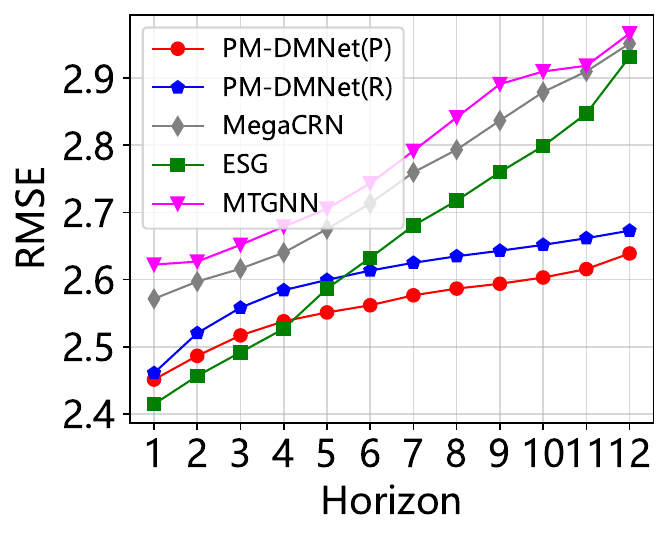}}
	\subfigure[{RMSE on NYC-Taxi16}]{
		\includegraphics[width=0.185\linewidth]{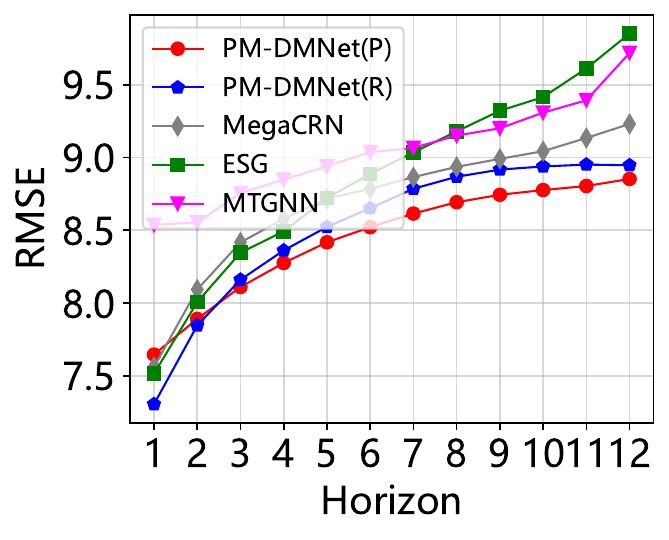}}	
        \subfigure[{RMSE on NYC-Bike14}]{
		\includegraphics[width=0.185\linewidth]{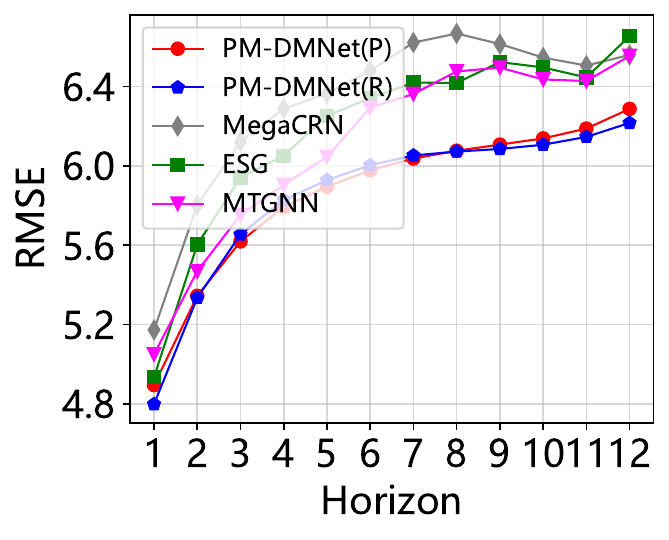}}	
        \subfigure[{RMSE on NYC-Bike15}]{
		\includegraphics[width=0.185\linewidth]{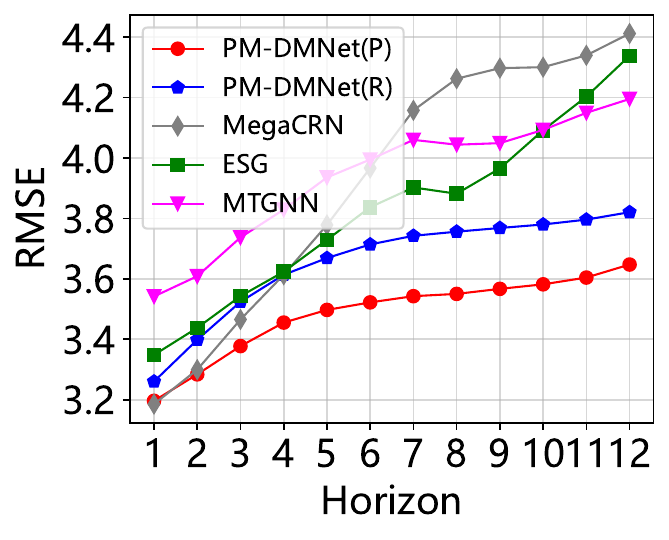}}	
        \subfigure[{RMSE on NYC-Taxi15}]{
		\includegraphics[width=0.1825\linewidth]{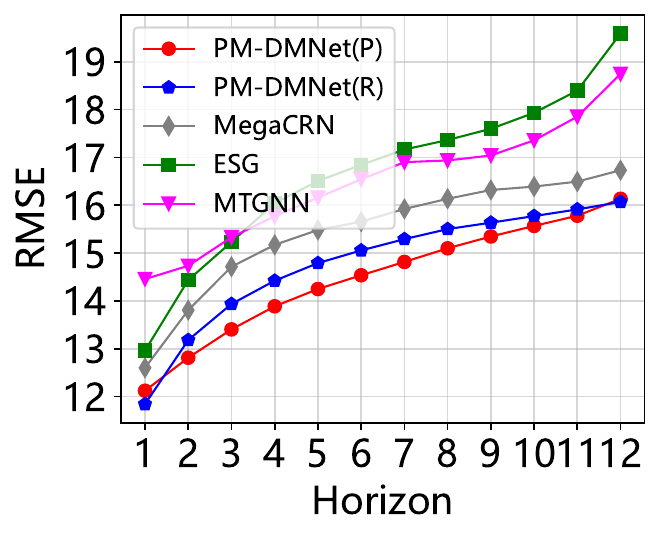}}

	\subfigure[{MAE on NYC-Bike16}]{
		\includegraphics[width=0.185\linewidth]{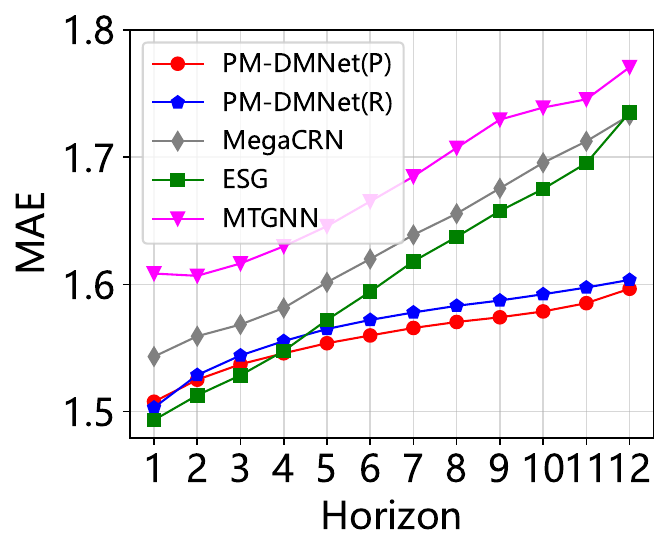}}
	\subfigure[{MAE on NYC-Taxi16}]{
		\includegraphics[width=0.185\linewidth]{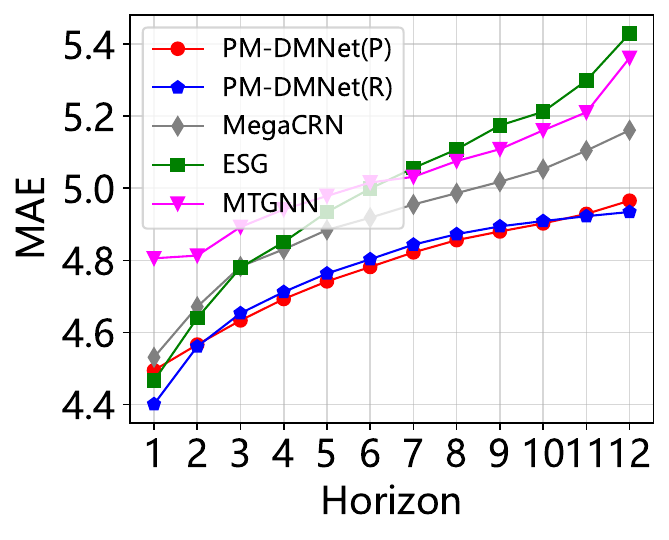}}	
        \subfigure[{MAE on NYC-Bike14}]{
		\includegraphics[width=0.185\linewidth]{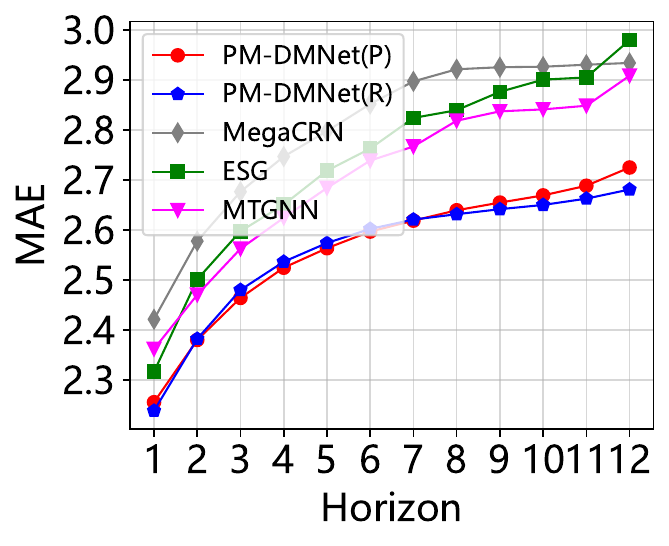}}	
        \subfigure[{MAE on NYC-Bike15}]{
		\includegraphics[width=0.185\linewidth]{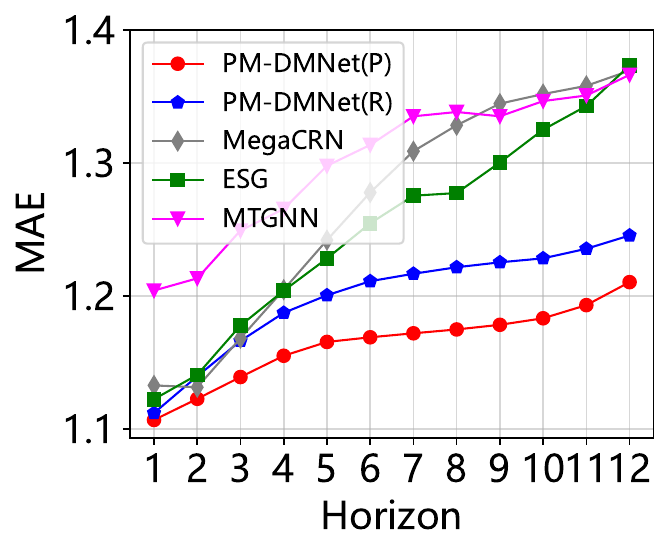}}	
        \subfigure[{MAE on NYC-Taxi15}]{
		\includegraphics[width=0.185\linewidth]{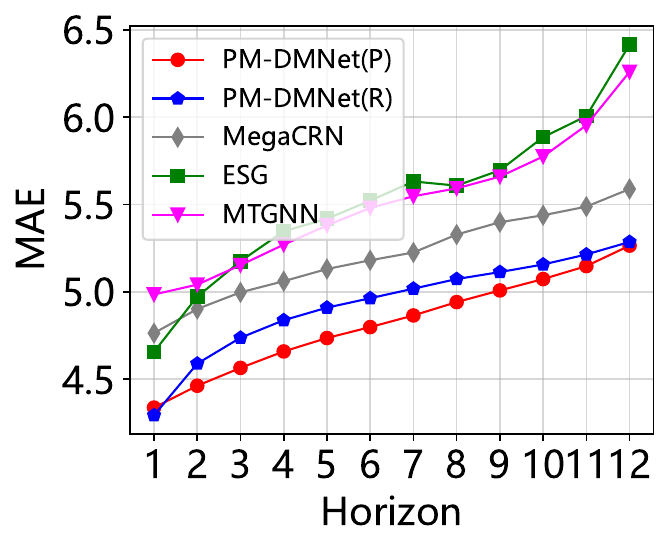}}
        \captionsetup{justification=centering}
	\caption{Prediction error at each horizon on five raffic demand datasets.}
        \label{fig: Prediction error at each horizon on five traffic demand datasets.}
\end{figure*}

\begin{table*}[ht]
\caption{Performance comparison between PM-DMNet and the baselines on five traffic flow/speed datasets. The best results are highlighted in bold, and the second-best results are underlined.}
\captionsetup{justification=centering}
\label{Performance comparison between PM-DMNet and the baselines on five traffic flow/speed datasets.}
 \resizebox{\linewidth}{!}{
\begin{tabular}{cccccccccccccccc}
\toprule[1pt]%
\multirow{2}{*}{Methods} & \multicolumn{3}{c}{PEMSD4}                         & \multicolumn{3}{c}{PEMSD7}                        & \multicolumn{3}{c}{PEMSD8}                        & \multicolumn{3}{c}{PEMSD7(M)}                   & \multicolumn{3}{c}{PEMSD7(L)}                   \\ \cline{2-16} 
                         & RMSE           & MAE            & MAPE             & RMSE           & MAE            & MAPE            & RMSE           & MAE            & MAPE            & RMSE          & MAE           & MAPE            & RMSE          & MAE           & MAPE            \\ \toprule[1pt]%
HA                       & 59.24          & 38.03          & 27.88\%          & 65.64          & 45.12          & 24.51\%         & 59.24          & 34.86          & 27.88\%         & 8.63          & 4.59          & 14.35\%         & 9.03          & 4.84          & 14.90\%         \\
ARIMA                    & 48.80          & 33.73          & 24.18\%          & 59.27          & 38.17          & 19.46\%         & 44.32          & 31.09          & 22.73\%         & 13.20         & 7.27          & 15.38\%         & 12.39         & 7.51          & 15.83\%         \\
VAR                      & 38.61          & 24.54          & 17.24\%          & 75.63          & 50.22          & 32.22\%         & 29.81          & 19.19          & 13.10\%         & 7.61          & 4.25          & 10.28\%         & 8.09          & 4.45          & 11.62\%         \\
SVR                      & 44.56          & 28.70          & 19.20\%          & 50.22          & 32.49          & 14.26\%         & 36.16          & 23.25          & 14.64\%         & 7.47          & 4.09          & 10.03\%         & 8.11          & 4.41          & 11.58\%         \\
LSTM                     & 40.65          & 26.77          & 18.23\%          & 45.94          & 29.98          & 13.20\%         & 35.17          & 23.09          & 14.99\%         & 7.51          & 4.16          & 10.10\%         & 8.20          & 4.66          & 11.69\%         \\
TCN                      & 37.26          & 23.22          & 15.59\%          & 42.23          & 32.72          & 14.26\%         & 35.79          & 22.72          & 14.03\%         & 7.20          & 4.36          & 9.71\%          & 7.29          & 4.05          & 10.43\%         \\
STGCN                    & 34.89          & 21.16          & 13.83\%          & 39.34          & 25.33          & 11.21\%         & 27.09          & 17.50          & 11.29\%         & 6.79          & 3.86          & 10.06\%         & 6.83          & 3.89          & 10.09\%         \\
DCRNN                    & 33.44          & 21.22          & 14.17\%          & 38.61          & 25.22          & 11.82\%         & 26.36          & 16.82          & 10.92\%         & 7.18          & 3.83          & 9.81\%          & 8.33          & 4.33          & 11.41\%         \\
GWN                      & 39.66          & 24.89          & 17.29\%          & 41.50          & 26.39          & 11.97\%         & 30.05          & 18.28          & 12.15\%         & 6.24          & 3.19          & 8.02\%          & 7.09          & 3.75          & 9.41\%          \\
ASTGCN(r)                & 35.22          & 22.93          & 16.56\%          & 37.87          & 24.01          & 10.73\%         & 28.06          & 18.25          & 11.64\%         & 6.18          & 3.14          & 8.12\%          & 6.81          & 3.51          & 9.24\%          \\
LSGCN                    & 33.86          & 21.53          & 13.18\%          & 41.46          & 27.31          & 11.98\%         & 26.76          & 17.73          & 11.20\%         & 5.98          & 3.05          & 7.62\%          & 6.55          & 3.49          & 8.77\%          \\
STSGCN                   & 33.65          & 21.19          & 13.90\%          & 39.03          & 24.26          & 10.21\%         & 26.80          & 17.13          & 10.96\%         & 5.93          & 3.01          & 7.55\%          & 6.88          & 3.61          & 9.13\%          \\
AGCRN                    & 32.26          & 19.83          & 12.97\%          & 36.55          & 22.37          & 9.12\%          & 25.22          & 15.95          & 10.09\%         & 5.84          & 2.99          & 7.42\%          & 6.04          & 3.13          & 7.75\%          \\
STFGNN                   & 32.51          & 20.48          & 16.77\%          & 36.60          & 23.46          & 9.21\%          & 26.25          & 16.94          & 10.60\%         & 5.74          & 2.93          & 7.28\%          & 5.96          & 3.07          & 7.71\%          \\
STGODE                   & 32.82          & 20.84          & 13.77\%          & 37.54          & 22.59          & 10.14\%         & 25.97          & 16.81          & 10.62\%         & 5.66          & 2.97          & 7.36\%          & 5.98          & 3.22          & 7.94\%          \\
STWave                   & 30.39          & 18.50          & 12.43\%          & 33.88          & 19.94          & 8.38\%          & 23.40          & {\ul 13.42}    & {\ul 8.90\%}    & 5.39          & 2.66          & 6.76\%          & 5.87          & 2.88          & 7.25\%          \\
MegaCRN                  & 31.03          & 19.07          & 12.71            & 33.83          & 20.42          & 8.68\%          & 24.15          & 15.19          & 9.88\%          & 5.40          & 2.67          & 6.73\%          & 5.84          & 2.88          & 7.19\%          \\ \toprule[1pt]%
PM-DMNet(P)              & \textbf{30.36} & \textbf{18.34} & {\ul 12.05\%}    & {\ul 33.33}    & {\ul 19.35}    & {\ul 8.05\%}    & {\ul 23.35}    & 13.55          & 9.04\%          & \textbf{5.33} & {\ul 2.61}    & \textbf{6.55\%} & \textbf{5.79} & {\ul 2.81}    & {\ul 7.13\%}    \\
PM-DMNet(R)              & {\ul 30.68}    & {\ul 18.37}    & \textbf{12.01\%} & \textbf{33.15} & \textbf{19.18} & \textbf{7.95\%} & \textbf{23.22} & \textbf{13.40} & \textbf{8.87\%} & {\ul 5.36}    & \textbf{2.60} & {\ul 6.57\%}    & {\ul 5.81}    & \textbf{2.79} & \textbf{6.99\%} \\ \toprule[1pt]%
\end{tabular}
}
\end{table*}

 \begin{figure*}[ht]
	\centering  
	\subfigbottomskip=2pt 
	\subfigcapskip=-5pt 
	\subfigure[{RMSE on PEMSD4}]{
		\includegraphics[width=0.185\linewidth]{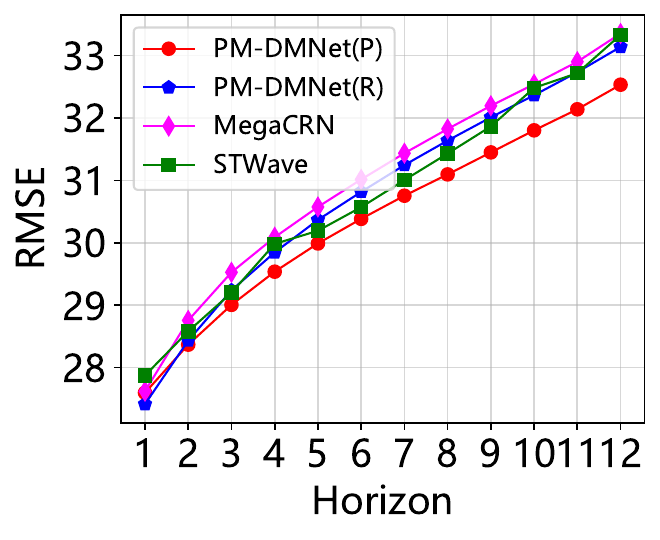}}
	\subfigure[{RMSE on PEMSD7}]{
		\includegraphics[width=0.185\linewidth]{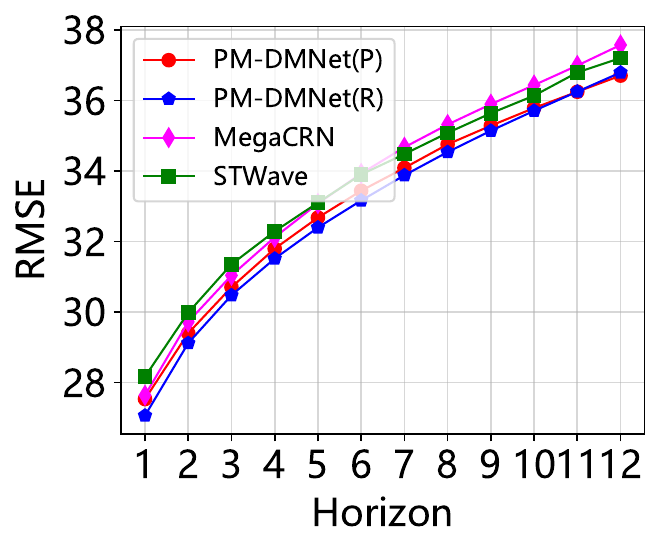}}	
        \subfigure[{RMSE on PEMSD8}]{
		\includegraphics[width=0.185\linewidth]{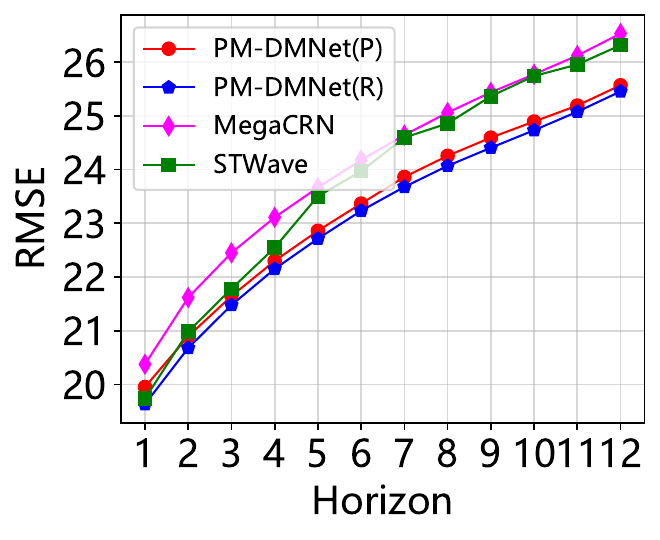}}	
        \subfigure[{RMSE on PEMSD7(M)}]{
		\includegraphics[width=0.18754\linewidth]{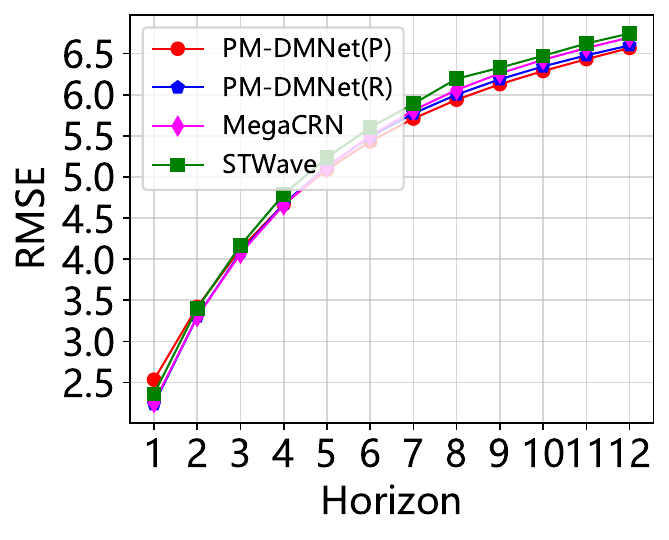}}	
        \subfigure[{RMSE on PEMSD7(L)}]{
		\includegraphics[width=0.18754\linewidth]{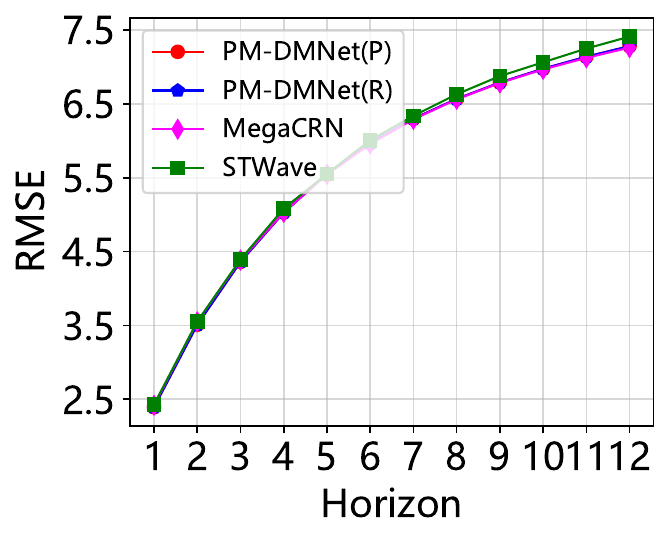}}

	\subfigure[{MAE on PEMSD4}]{
		\includegraphics[width=0.185\linewidth]{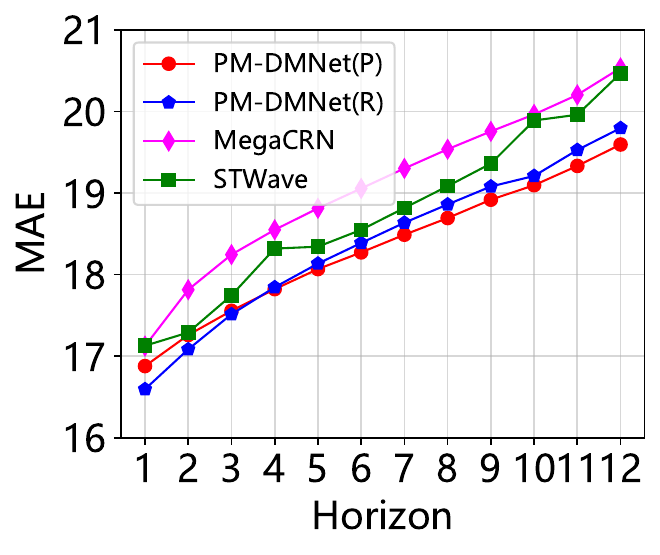}}
	\subfigure[{MAE on PEMSD7}]{
		\includegraphics[width=0.185\linewidth]{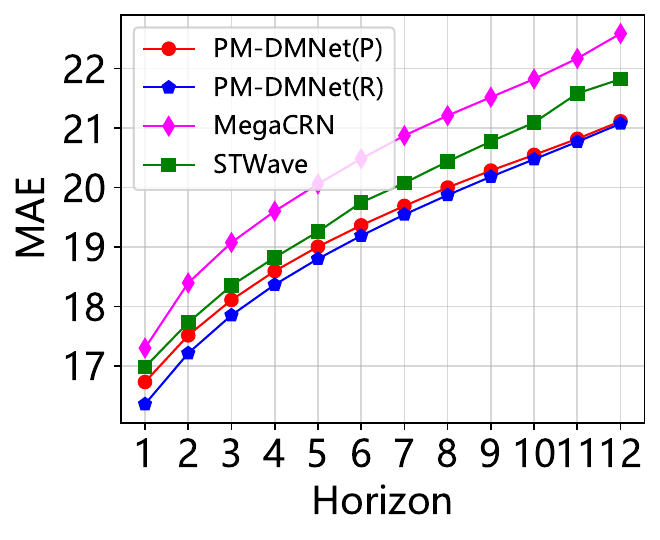}}	
        \subfigure[{MAE on PEMSD8}]{
		\includegraphics[width=0.185\linewidth]{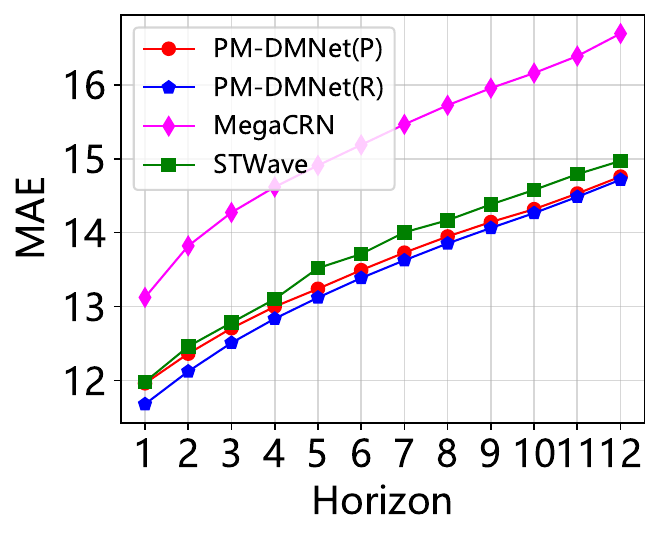}}	
        \subfigure[{MAE on PEMSD7(M)}]{
		\includegraphics[width=0.18754\linewidth]{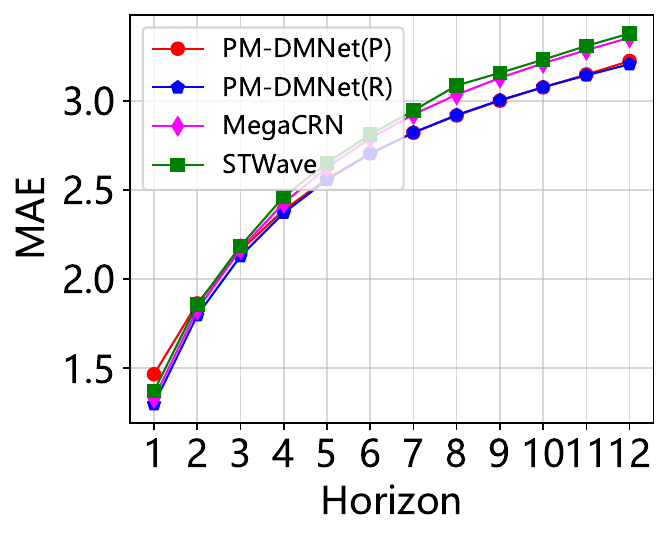}}	
        \subfigure[{MAE on PEMSD7(L)}]{
		\includegraphics[width=0.18754\linewidth]{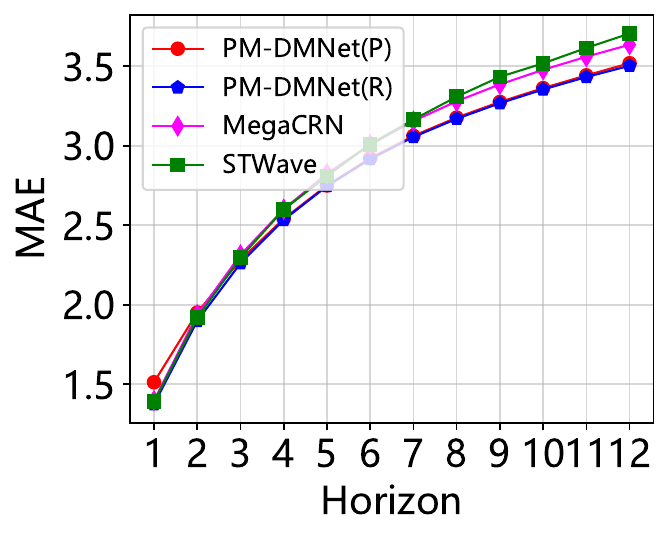}}
        \captionsetup{justification=centering}
	\caption{Prediction error at each horizon on five flow/speed datasets.}
        \label{fig: Prediction error at each horizon on five flow/speed datasets.}
\end{figure*}

\subsection{Metrics}
The following three evaluation metrics are chosen to assess model performance: Root Mean Square Error (RMSE), Mean Absolute Error (MAE), Mean Absolute Percentage Error (MAPE) and Empirical Correlation Coefficient (CORR).

\begin{equation}
    MAE=\frac{1}{\phi}\sum_{i=1}^{\phi}{|Y_i-\hat{Y}_i|} \label{MAE}
\end{equation}

\begin{equation}
    RMSE=\sqrt{\frac{1}{\phi}\sum_{i=1}^{\phi}{(Y_i-\hat{Y}_i )^2}} \label{RMSE}
\end{equation}

\begin{equation}
    MAPE=\frac{1}{\phi}\sum_{i=1}^{\phi}{|\frac{Y_i-\hat{Y}_i}{Y_i}|} \label{MAPE}
\end{equation}

\begin{equation}
    CORR=\frac{1}{N}
    \sum_{n=1}^{N}{\frac{\sum_{i=1}^{\phi}(Y_{n,i}-\overline{Y}_{n})(\hat{Y}_{n,i}-\overline{\hat{Y}}_{n})}{\sqrt{\sum_{i=1}^{\phi}(Y_{n,i}-\overline{Y}_{n})^{2}(\hat{Y}_{n,i}-\overline{\hat{Y}}_{n})^{2}}}} \label{CORR}
\end{equation}
where $\phi$ represents the length of the predicted sequence, and $\overline{Y}_{n}$ and $\overline{\hat{Y}}_{n}$ denote the mean values of the true and predicted values at node $n$, respectively.
A smaller value of these metrics indicates higher prediction accuracy and better prediction performance.

\section{Experiments}
\subsection{Performance Comparison}
Table \ref{Performance comparison between PM-DMNet and the baselines on five traffic demand datasets.} presents the results of our model and baselines across different datasets. Clearly, optimal results are achieved by our model across all five datasets. XGBoost, being a machine learning model, fails to capture the nonlinear relationships within traffic condition, resulting in its inferior performance. DCRNN, STGCN, and STG2Seq utilize predefined graph structures to capture spatio-temporal correlations within traffic condition, yielding satisfactory outcomes. However, due to the fixed weights in these predefined graph structures, the inability to capture dynamic correlations leaves significant room for improvement. MTGNN and GTS demonstrate commendable progress by learning graph structures adaptively from the data. Nevertheless, these adaptive graphs remain static and fail to capture the dynamic relationships between nodes. MegaCRN employs a meta-graph learner to construct dynamic graphs for extracting correlations between nodes. However, it does not consider the influence of temporal information on traffic patterns, which limits its performance. PM-DMNet excels by leveraging a dynamic memory network to dynamically extract features by identifying the most analogous traffic patterns based on historical data. 
Figure \ref{fig: Prediction error at each horizon on five traffic demand datasets.} illustrates the prediction errors of PM-DMNet compared to three baseline models across different prediction horizons. It is observed that, except for the initial three prediction steps, PM-DMNet consistently achieves lower prediction errors than the baseline models. Additionally, the error growth rate of PM-DMNet across all time horizons is slower than that of the baseline models. Benefiting from the functionality of the evolving graph, ESG achieves comparable short-term prediction performance to PM-DMNet. However, as the prediction horizon expands, the error growth rate of ESG becomes significantly faster than that of PM-DMNet, resulting in an overall performance inferior to PM-DMNet. By leveraging temporal information corresponding to the prediction targets, PM-DMNet substantially reduces prediction uncertainty, thereby enhancing performance.

Table \ref{Performance comparison between PM-DMNet and the baselines on five traffic flow/speed datasets.} presents the results of our model and baseline models on traffic flow/speed datasets. It is observed that, except for PEMSD8 where STWave slightly outperforms PM-DMNet (P) and is comparable to PM-DMNet (R), our model achieves the best performance across all datasets. Figure \ref{fig: Prediction error at each horizon on five flow/speed datasets.} shows the prediction errors of PM-DMNet and the two other best baseline models at different prediction horizons. From Figure \ref{fig: Prediction error at each horizon on five flow/speed datasets.}, it is evident that the error gaps between models are more pronounced in the flow datasets compared to the speed datasets, indicating that predicting traffic speed is more challenging than predicting traffic flow.
STWave utilizes the DWT algorithm to decompose traffic data into two separate low-frequency and high-frequency sequences, modeling them independently while considering the impact of temporal information, resulting in good performance on traffic flow datasets. However, on speed datasets, due to the inherent differences between traffic speed and traffic flow, the DWT algorithm struggles to decompose useful high and low-frequency sequences, causing STWave's performance to be on par with MegaCRN. PM-DMNet does not rely on sequence decomposition for modeling, thus avoiding the difficulties associated with ineffective decomposition, leading to excellent performance on both flow and speed datasets.

\begin{table}[ht]
\caption{The computation cost on four datasets.}
\captionsetup{justification=centering}
\label{The computation cost on four datasets.}
\resizebox{\linewidth}{!}{
\begin{tabular}{ccccc}
\toprule[1pt]%
Dataset                 & Model       & \begin{tabular}[c]{@{}c@{}}Tainning Time\\ (s/epoch)\end{tabular} & \begin{tabular}[c]{@{}c@{}}Inference Time\\ (s)\end{tabular} & \begin{tabular}[c]{@{}c@{}}GPU Cost\\ (GB)\end{tabular} \\ \toprule[1pt]%
\multirow{4}{*}{NYC-Bike16}   & PM-DMNet(P) & 4.17                                                              & 0.29                                                         & 1.44                                                    \\
                        & PM-DMNet(R) & 7.26                                                              & 0.47                                                         & 1.46                                                    \\
                        & ESG         & 20.83                                                             & 1.65                                                         & 15.60                                                   \\
                        & MegaCRN     & 7.04                                                              & 0.68                                                         & 2.00                                                    \\ \toprule[1pt]%
\multirow{4}{*}{NYC-Taxi16}   & PM-DMNet(P) & 4.43                                                              & 0.29                                                         & 1.50                                                    \\
                        & PM-DMNet(R) & 7.53                                                              & 0.46                                                         & 1.50                                                    \\
                        & ESG         & 22.93                                                             & 1.91                                                         & 16.50                                                   \\
                        & MegaCRN     & 6.60                                                              & 0.66                                                         & 2.23                                                    \\ \toprule[1pt]%
\multirow{4}{*}{PEMSD4} & PM-DMNet(P) & 14.87                                                             & 1.53                                                         & 1.73                                                    \\
                        & PM-DMNet(R) & 25.21                                                             & 2.35                                                         & 1.70                                                    \\
                        & STWave      & 56.32                                                             & 7.46                                                         & 5.79                                                    \\
                        & MegaCRN     & 24.60                                                             & 3.71                                                         & 2.44                                                    \\ \toprule[1pt]%
\multirow{4}{*}{PEMSD7} & PM-DMNet(P) & 33.77                                                             & 4.08                                                         & 4.79                                                    \\
                        & PM-DMNet(R) & 41.68                                                             & 4.09                                                         & 4.75                                                    \\
                        & STWave      & 272.95                                                            & 36.53                                                        & 16.76                                                   \\
                        & MegaCRN     & 104.12                                                            & 16.91                                                        & 7.57                                                    \\ \toprule[1pt]%
\end{tabular}
}
\end{table}

\subsection{Computation Cost}


To compare and demonstrate the computational efficiency of our model, we evaluate the training time, inference time, and GPU cost of selected models. The batch size for all models is set to 32. Table \ref{The computation cost on four datasets.} shows the computational costs of PM-DMNet compared to baseline models. As observed in Table \ref{The computation cost on four datasets.}, the training and inference times of PM-DMNet(P) are significantly lower than those of other baselines, and it also outperforms PM-DMNet(R), demonstrating the advantages of the dynamic memory network and PMP in terms of computational speed and memory usage.
Despite ESG's strong performance, its high GPU cost and relatively slow processing speed present challenges in deployment. Although STWave employs a novel graph attention mechanism to optimize modeling speed, its complex network structure still demands substantial GPU resources and long training times. MegaCRN uses RMP methods while adopting a simple adaptive graph convolution method to extract spatial correlations between nodes, resulting in lower training time and GPU cost. Therefore, on datasets with fewer nodes, MegaCRN's training time is comparable to that of PM-DMNet(P), which also uses RMP methods. However, on large-scale node datasets, the $O(N^2)$ complexity of GCN still requires higher training time and GPU cost. In contrast, the dynamic memory network used by PM-DMNet(P) has a time complexity of $O(N)$, significantly lower than the $O(N^2)$ complexity of graph convolution networks (GCNs). Consequently, on PEMSD7, PM-DMNet(P) exhibits faster training speed and lower GPU cost than MegaCRN, showcasing the computational speed advantages of our model.

\subsection{Complexity Analysis}

The computation complexity for feature aggregation in GCN is $O(N^2)$, and both the computation of attention matrices and feature aggregation in attention mechanisms are also $O(N^2)$. For DMN, the computation complexity for calculating similarity weights and feature aggregation is $O(NM)$, where $M$ is a constant value significantly smaller than $N$. When $M$ is much smaller than $N$, the time complexity of DMN can be considered as $ O(N)$. Therefore, compared to GCN and attention mechanisms, DMN exhibits notable advantages in terms of time and memory complexity.

 \begin{figure*}[ht]
	\centering  
	\subfigbottomskip=2pt 
	\subfigcapskip=-5pt 
        \captionsetup[subfigure]{justification=centering}
	\subfigure[{RMSE on NYC-Bike16}]{
		\includegraphics[width=0.23\linewidth]{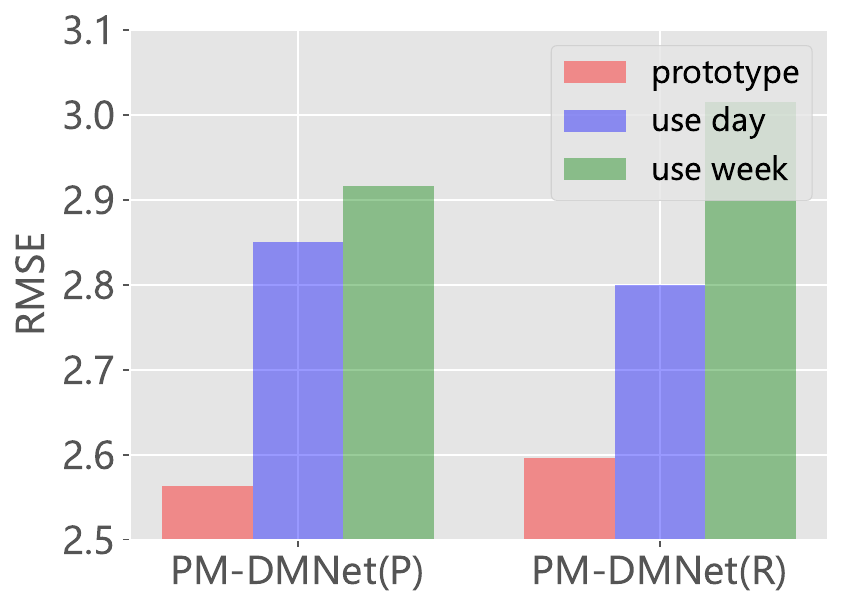}}
        \subfigure[{RMSE on NYC-Taxi16}]{
		\includegraphics[width=0.23596\linewidth]{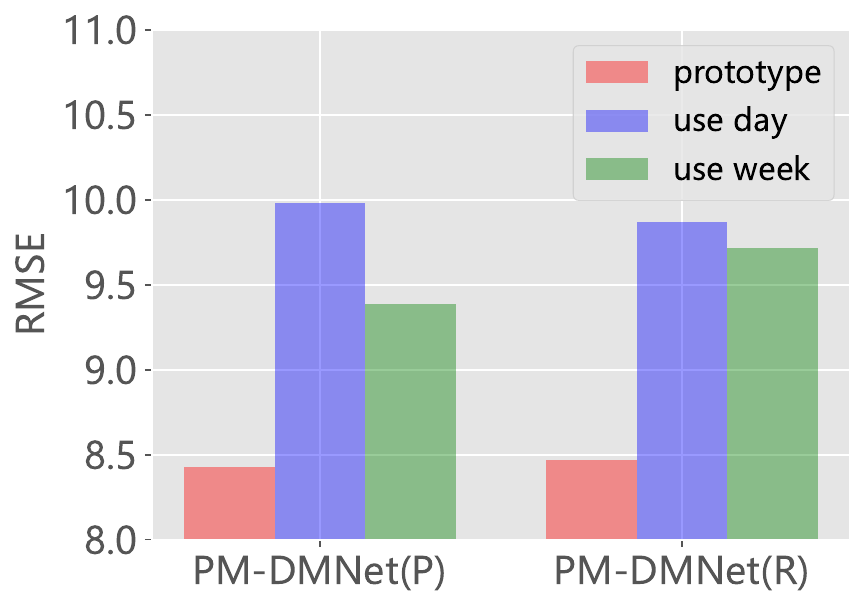}}	
        \subfigure[{RMSE on PEMSD4}]{
		\includegraphics[width=0.22755\linewidth]{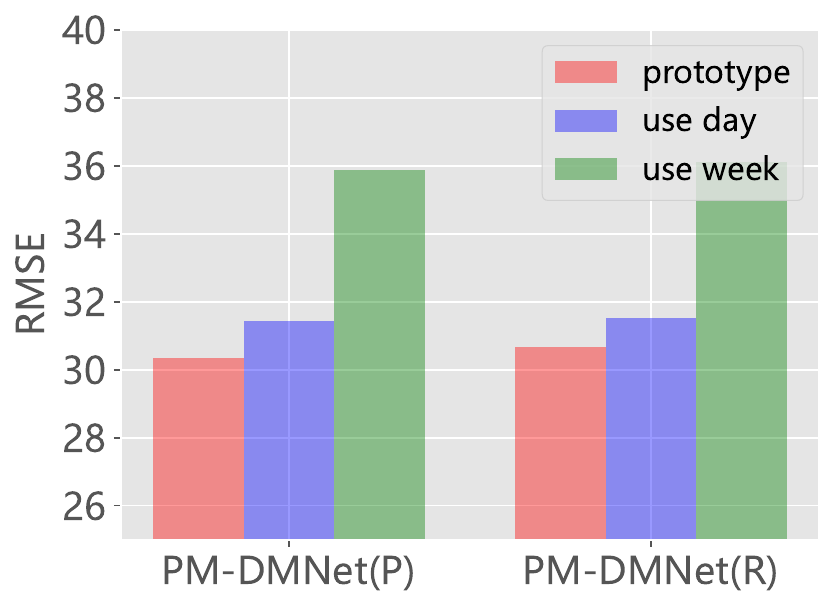}}	
        \subfigure[{RMSE on PEMSD7(M)}]{
		\includegraphics[width=0.23\linewidth]{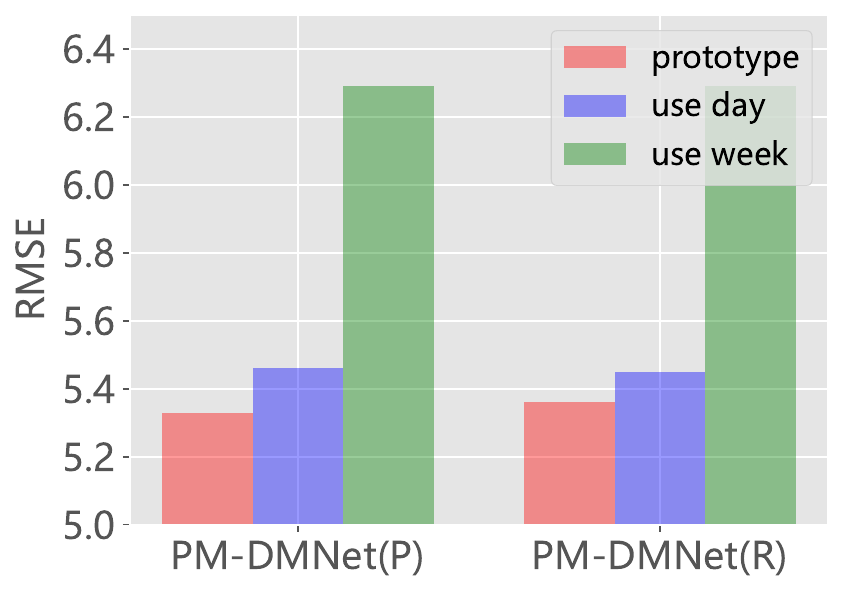}}

	\subfigure[{MAE on NYC-Bike16}]{
		\includegraphics[width=0.23\linewidth]{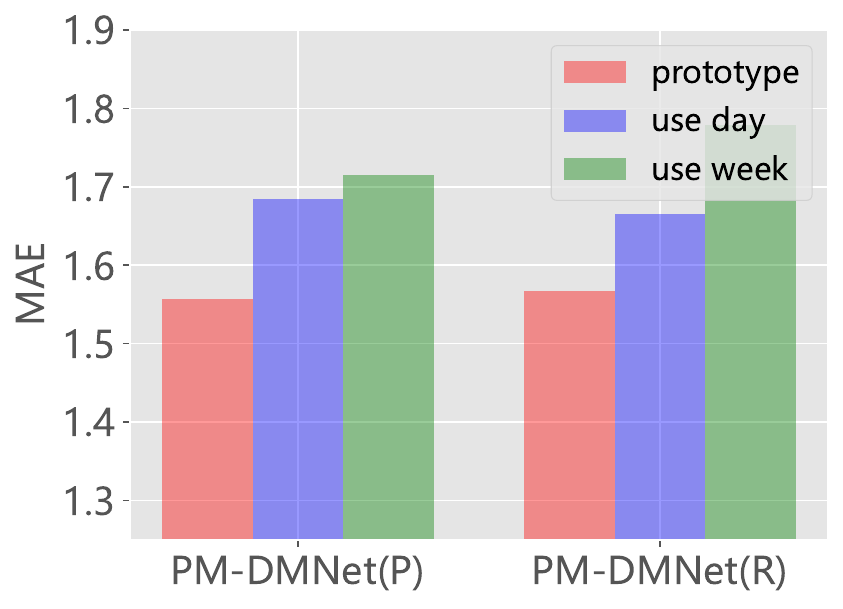}}
        \subfigure[{MAE on NYC-Taxi16}]{
		\includegraphics[width=0.23596\linewidth]{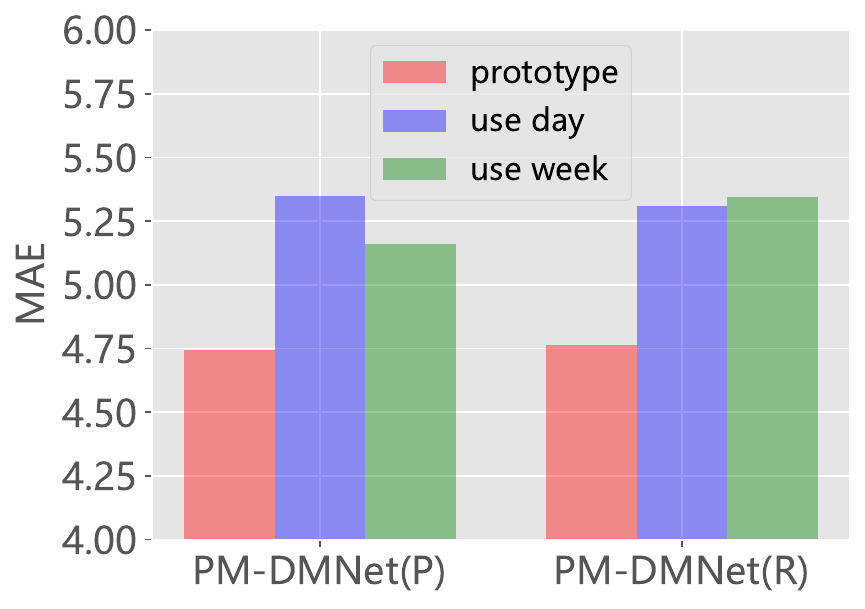}}	
        \subfigure[{MAE on PEMSD4}]{
		\includegraphics[width=0.22755\linewidth]{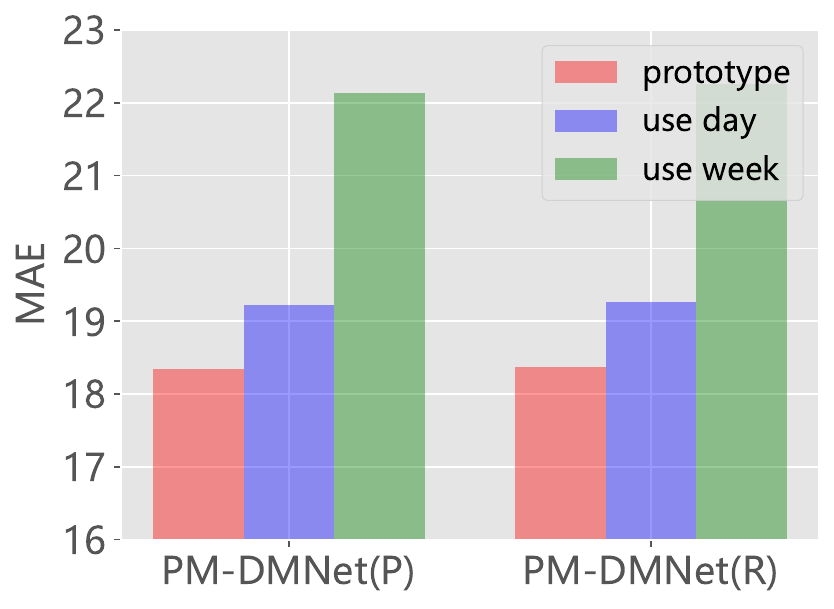}}	
        \subfigure[{MAE on PEMSD7(M)}]{
		\includegraphics[width=0.23\linewidth]{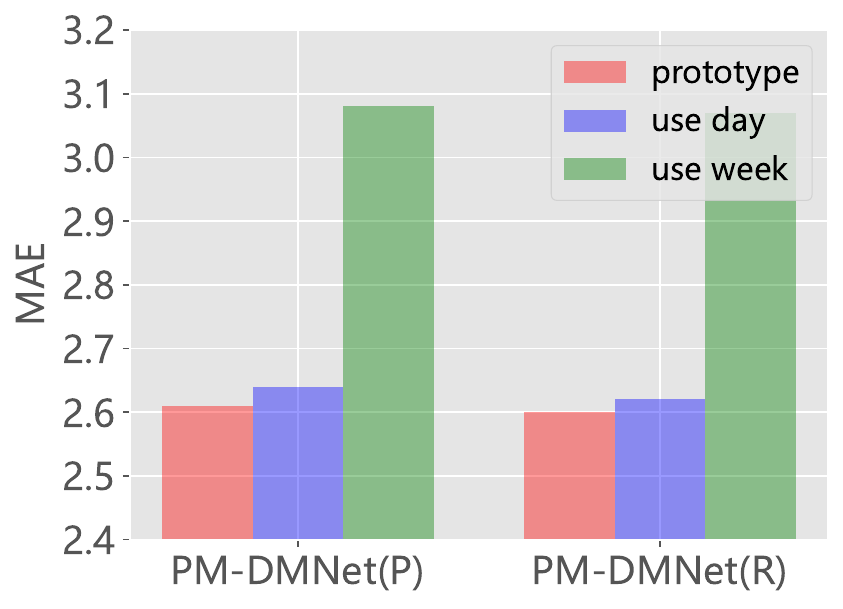}}
        \captionsetup{justification=centering}
	\caption{Ablation experiment of time embedding.}
        \label{Ablation experiment of time embedding.}
\end{figure*}

\subsection{Recurrent Multi-step Prediction vs. Parallel Multi-step Prediction}

In this subsection, the performance of PMP and RMP prediction methods is compared. Tables \ref{Performance comparison between PM-DMNet and the baselines on five traffic demand datasets.} and \ref{Performance comparison between PM-DMNet and the baselines on five traffic flow/speed datasets.} present the results of PM-DMNet(P) and PM-DMNet(R) on traffic demand and traffic flow/speed datasets, respectively. As shown in Table \ref{Performance comparison between PM-DMNet and the baselines on five traffic demand datasets.}, PM-DMNet(P) outperforms PM-DMNet(R) on three datasets for traffic demand prediction tasks and matches PM-DMNet(R) on two datasets, indicating that PM-DMNet(P) has certain advantages over PM-DMNet(R) in traffic demand prediction tasks. This is further evidenced by the per-step prediction errors shown in Figure \ref{fig: Prediction error at each horizon on five traffic demand datasets.}.

However, as seen in Table \ref{Performance comparison between PM-DMNet and the baselines on five traffic flow/speed datasets.}, PM-DMNet(R) exhibits a performance advantage over PM-DMNet(P) in traffic flow/speed tasks. In Figure \ref{fig: Prediction error at each horizon on five flow/speed datasets.}, it is shown that PM-DMNet(P) has significantly larger prediction errors in the initial time steps compared to PM-DMNet(R), but the errors of both methods are comparable in the later time steps. This phenomenon might be attributed to the different time intervals of the datasets. The traffic demand datasets are collected at 30-minute intervals, resulting in more pronounced differences between historical data and prediction targets, where the PMP method performs better than the RMP method. In contrast, traffic flow/speed datasets are collected at 5-minute intervals, creating more continuity between historical data and prediction targets, thereby giving the RMP method an edge over the PMP method in performance.


\section{Ablation Study}
\subsection{Effectiveness Analysis of model components}
In this section, ablation experiments are conducted on the key components of PM-DMNet to validate their effectiveness. To investigate the impact of different modules, the following variants are designed:

\textbf{W/O Decoder}: This variant removes the decoder component and predicts using an MLP layer directly applied to the encoder's output. Since the decoding process is omitted, this variant is identical for both PM-DMNet(P) and PM-DMNet(R).

\textbf{W/O TAM}: In this variant, the Transfer Attention Module (TAM) is excluded. Instead, the prediction is made using the output $H_n$ from the encoder, replacing the output $H_{n+1}$ from the transfer attention mechanism.

\textbf{W/O DMN}: This variant substitutes the Dynamic Memory Network (DMN) module with an MLP layer for making predictions.

\textbf{W/O NAPL}: This variant removes the Node Adaptive Parameter learning (NAPL) module and uses a linear layer instead.


\begin{table}[ht]
\caption{Ablation experiments for each module.}
\captionsetup{justification=centering}
\label{Ablation experiments for each module.}
\resizebox{\linewidth}{!}{
\begin{tabular}{cccccccc}
\toprule[1pt]%
\multirow{2}{*}{dataset} & \multirow{2}{*}{variants}        & \multicolumn{3}{c}{PM-DMNet(P)}                                           & \multicolumn{3}{c}{PM-DMNet(R)}                      \\ \cline{3-8} 
                         &                                  & RMSE            & MAE             & CORR                                  & RMSE            & MAE             & CORR             \\ \toprule[1pt]%
\multirow{5}{*}{NYC-Bike16		}    & \multicolumn{1}{c|}{PM-DMNet}    & \textbf{2.5631} & \textbf{1.5566} & \multicolumn{1}{c|}{\textbf{0.7709}}  & \textbf{2.5964} & \textbf{1.5667} & \textbf{0.7638}  \\
                         & \multicolumn{1}{c|}{\textbf{W/O Decoder}} & 2.6308          & 1.5949          & \multicolumn{1}{c|}{0.7602}           & 2.6308          & 1.5949          & 0.7602           \\
                         & \multicolumn{1}{c|}{\textbf{W/O TAM}}     & 2.6341          & 1.5859          & \multicolumn{1}{c|}{0.7599}           & /               & /               & /                \\
                         & \multicolumn{1}{c|}{\textbf{W/O DMN}}     & 3.1756          & 1.8078          & \multicolumn{1}{c|}{0.6728}           & 3.9676          & 2.2438          & 0.4815           \\
                         & \multicolumn{1}{c|}{\textbf{W/O NAPL}}      & 3.1800          & 1.8057          & \multicolumn{1}{c|}{0.6726}           & 3.2525          & 18.265          & 0.6689           \\ \toprule[1pt]%
\multirow{2}{*}{dataset} & \multirow{2}{*}{variants}        & \multicolumn{3}{c}{PM-DMNet(P)}                                           & \multicolumn{3}{c}{PM-DMNet(R)}                      \\ \cline{3-8} 
                         &                                  & RMSE            & MAE             & MAPE                                  & RMSE            & MAE             & MAPE             \\ \toprule[1pt]%
\multirow{5}{*}{PEMSD4}  & \multicolumn{1}{c|}{PM-DMNet}    & \textbf{30.36}  & \textbf{18.34}  & \multicolumn{1}{c|}{\textbf{12.05\%}} & \textbf{30.68}  & \textbf{18.37}  & \textbf{12.01\%} \\
                         & \multicolumn{1}{c|}{\textbf{W/O Decoder}} & 33.31           & 20.15           & \multicolumn{1}{c|}{13.28\%}          & 33.31           & 20.15           & 13.28\%          \\
                         & \multicolumn{1}{c|}{\textbf{W/O TAM}}     & 30.75           & 18.42           & \multicolumn{1}{c|}{12.10\%}          & /               & /               & /                \\
                         & \multicolumn{1}{c|}{\textbf{W/O DMN}}     & 35.03           & 21.40           & \multicolumn{1}{c|}{14.29\%}          & 39.74           & 25.32           & 17.22\%          \\
                         & \multicolumn{1}{c|}{\textbf{W/O NAPL}}      & 34.84           & 21.19           & \multicolumn{1}{c|}{14.31\%}          & 34.98           & 21.3            & 14.29\%          \\ \toprule[1pt]%
\end{tabular}
}
\end{table}

Table \ref{Ablation experiments for each module.} presents the performance of PM-DMNet(P) and PM-DMNet(R) alongside their variants. It is evident from the table that PM-DMNet(P) and PM-DMNet(R) outperform all other variants, demonstrating the effectiveness of each component.

For the \textbf{W/O Decoder} variant, the pattern matching process is omitted during the decoding stage, and predictions are made directly using an MLP layer. As a result, this variant can only utilize historical data information and lacks the ability to leverage the time point information of the prediction target. Consequently, its performance is inferior to PM-DMNet(P) and PM-DMNet(R).

The performance of the \textbf{W/O TAM} variant also falls short of PM-DMNet(P). This indicates that the discrepancy between historical data and the prediction target leads to a performance decline, validating our proposed solution. This shows that using a suitable method for parallel prediction can achieve results comparable to or better than serial prediction.

The \textbf{W/O DMN} variant's performance is significantly inferior to both PM-DMNet models, highlighting the feasibility of our approach to use a memory network to match and extract the most representative traffic patterns.

Similarly, the performance of the \textbf{W/O NAPL} variant is lower than that of the two PM-DMNet models, underscoring the necessity for the model to learn the unique traffic patterns of each node.




\subsection{Effectiveness Analysis of GCN and DMN}

To validate the differences in performance and computational cost between GCN and DMN, a variant named DGCNet is designed. This variant uses dynamic graph convolution instead of DMN. The formula for dynamic graph convolution is expressed as follows:
\begin{equation}
 E^{d}_{t}= F_{t}\odot T_{t}
\end{equation}
\begin{equation}
\begin{aligned}
A^{d}_{t}=ReLU(E^{d}_{t}{E^{d}_{t}}^{T})
 \end{aligned}
\end{equation}
\begin{equation}
  H_{t}=(I_{N}+D^{-\frac{1}{2}}A^{d}_{t}D^{-\frac{1}{2}})X\Theta
\end{equation}
where $A^{d}_{t}\in R^{N \times N}$ represents the dynamic graph at time point \( t \), $D$ is the degree matrix of $A^{d}_{t}$, and \( I_N \) represents the identity matrix. Similar to PM-DMNet, DGCNet can be divided into two variants based on the prediction method: DGCNet(P) and DGCNet(R).

\begin{table}[ht]
\caption{Ablation experiment of GCN and DMN}
\label{Ablation experiment of GCN and DMN}
\resizebox{\linewidth}{!}{
\begin{tabular}{cccccccc}
\toprule[1pt]%
Dataset                                                                & model       & RMSE           & MAE            & MPAE            & \begin{tabular}[c]{@{}c@{}}Train Time\\ (s/epoch)\end{tabular} & \begin{tabular}[c]{@{}c@{}}Inference Time\\  (s)\end{tabular} & \begin{tabular}[c]{@{}c@{}}GPU Cost\\ (GB)\end{tabular} \\ \toprule[1pt]%
\multirow{4}{*}{\begin{tabular}[c]{@{}c@{}}PEMSD7\\ (16)\end{tabular}} & DGCNet(P)   & 33.81          & 19.60          & 8.28\%          & 231.10                                                         & 23.62                                                         & 13.04                                                   \\
                                                                       & PM-DMNet(P) & \textbf{23.35} & \textbf{13.55} & \textbf{9.04\%} & \textbf{49.16}                                                 & \textbf{5.68}                                                 & \textbf{2.96}                                           \\ \cline{2-8} 
                                                                       & DGCNet(R)   & 33.38          & 19.39          & 8.06\%          & 237.98                                                         & 24.00                                                         & 12.58                                                   \\
                                                                       & PM-DMNet(R) & \textbf{23.22} & \textbf{13.40} & \textbf{8.87\%} & \textbf{81.43}                                                 & \textbf{7.67}                                                 & \textbf{3.45}                                           \\ \toprule[1pt]%
\multirow{4}{*}{\begin{tabular}[c]{@{}c@{}}PEMSD8\\ (64)\end{tabular}} & DGCNet(P)   & 23.99          & 13.95          & 9.29\%          & 7.99                                                           & 1.00                                                          & 3.20                                                    \\
                                                                       & PM-DMNet(P) & \textbf{33.33} & \textbf{19.35} & \textbf{8.05\%} & \textbf{7.82}                                                  & \textbf{0.84}                                                 & \textbf{1.64}                                           \\ \cline{2-8} 
                                                                       & DGCNet(R)   & 23.55          & 13.70          & 8.92\%          & 13.56                                                          & 1.32                                                          & 2.87                                                    \\
                                                                       & PM-DMNet(R) & \textbf{33.15} & \textbf{19.18} & \textbf{7.95\%} & \textbf{13.48}                                                 & \textbf{1.26}                                                 & \textbf{1.49}                                           \\ \toprule[1pt]%
\end{tabular}
}
\end{table}

Experiments are conducted on PEMSD7 and PEMSD8, with a batch size of 16 for PEMSD7 and 64 for PEMSD8. Table \ref{Ablation experiment of GCN and DMN} presents the results of GCN and DMN on these two datasets. It can be observed that PM-DMNet outperforms DGCNet, indicating that DMN can achieve excellent performance without relying on GCN. Additionally, while PM-DMNet's computational metrics are slightly better than DGCNet on the smaller PEMSD8 dataset, the difference is not significant. However, on the larger PEMSD7 dataset, PM-DMNet's computational metrics are significantly superior to those of DGCNet, demonstrating the advantage of DMN's \(O(N)\) complexity over GCN's \(O(N^2)\) complexity in large-scale node scenarios.

\subsection{Effectiveness Analysis of time embedding}
To validate the impact of intra-daily time features and weekly time features on the model, two variants are designed for this subsection:

\textbf{use day}: The dynamic memory network is updated using only intra-daily time feature embeddings in this variant.

\textbf{use week}: The dynamic memory network is updated using only weekly time feature embeddings in this variant.

Experiments are conducted on four datasets to observe the influence of time information on model performance across different types of data. 

Figure \ref{Ablation experiment of time embedding.} presents the performance of PM-DMNet(P) and PM-DMNet(R) along with their variants. 
It can be observed that when only one type of time feature embedding is used, the model's performance generally decreases. Except for the NYC-Taxi16 dataset, where \textbf{use week} outperforms \textbf{use day} in PM-DMNet(P), the performance of \textbf{use day} is superior to \textbf{use week} in all other cases. This indicates that intra-daily information typically has a greater impact on model performance than weekly information. Additionally, in the PEMSD7(M) dataset, the performance of \textbf{use day} is comparable to that of their original models, while the performance of \textbf{use week} varies significantly. This suggests that, unlike other types of data, traffic speed data shows less pronounced differences between weekdays and weekends, exhibiting high similarity.


\section{hyper-parameter analysis}

To validate the impact of hyperparameters on model performance, hyperparameter experiments are conducted on the PEMSD8 dataset. Specifically, in this section, we investigate the effects of the temporal embedding dimension \( p \), the dimension \( d \) of the node embedding matrix \( E \) in the node adaptive module, and the dimension \( M \) of the memory network matrix. In these experiments, other parameters are kept constant while only the parameter under study is changed.

 \begin{figure}[ht]
	\centering  
	\subfigbottomskip=2pt 
	\subfigcapskip=-5pt 
	\subfigure[{Sensitivity of Parameter $p$ to PM-DMNet(P)}]{
        \label{Sensitivity of Parameter $p$ to PM-DMNet(P)}
		\includegraphics[width=0.48\linewidth]{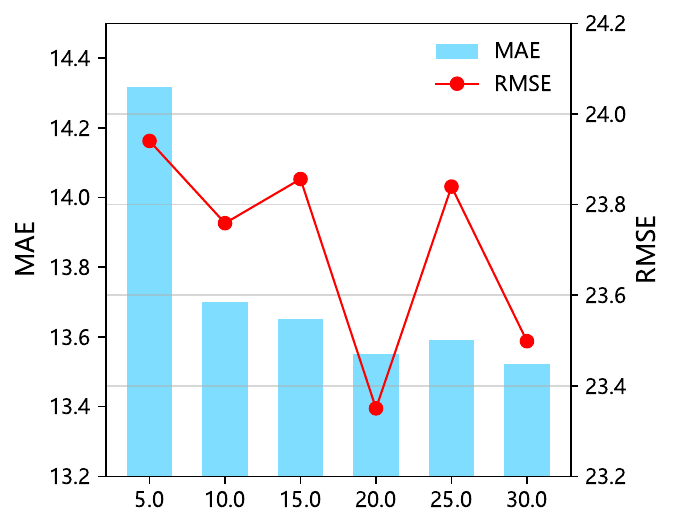}}
	\subfigure[{Sensitivity of Parameter $p$ to PM-DMNet(R)}]{
        \label{Sensitivity of Parameter $p$ to PM-DMNet(R)}
		\includegraphics[width=0.48\linewidth]{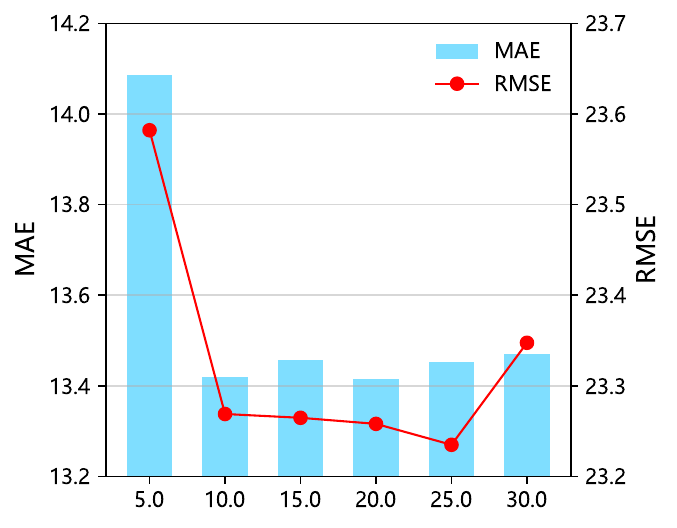}}
	\caption{Sensitivity analysis of parameter $p$ on PEMSD8.}
   \label{Sensitivity analysis of parameter $p$ on PEMSD8.}
\end{figure}

\subsection{Sensitivity to $p$}
The parameter \( p \) is set to \{5, 10, 15, 20, 25, 30\} to evaluate its sensitivity on model performance. In Figure \ref{Sensitivity analysis of parameter $p$ on PEMSD8.}, the performance of the model under different values of \( p \) is shown. It can be seen that the model performs relatively stable when \( p \) is between 10 and 25. Additionally, across various settings, PM-DMNet(R) consistently exhibits lower errors compared to PM-DMNet(P).

 \begin{figure}[ht]
	\centering  
	\subfigbottomskip=2pt 
	\subfigcapskip=-5pt 
	\subfigure[{Sensitivity of Parameter $d$ to PM-DMNet(P)}]{
        \label{Sensitivity of Parameter $d$ to PM-DMNet(P)}
		\includegraphics[width=0.48\linewidth]{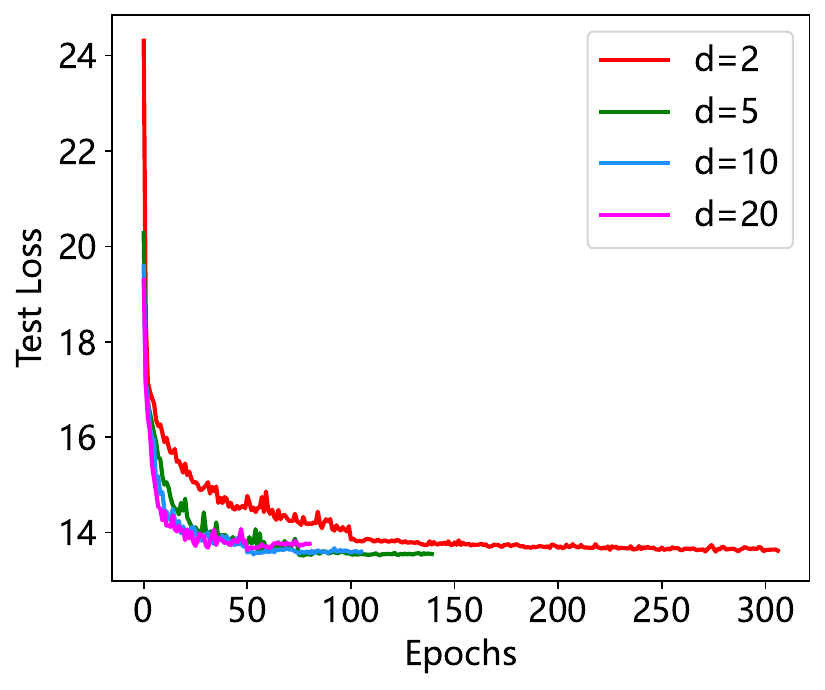}}
	\subfigure[{Sensitivity of Parameter $d$ to PM-DMNet(R)}]{
        \label{Sensitivity of Parameter $d$ to PM-DMNet(R)}
		\includegraphics[width=0.48\linewidth]{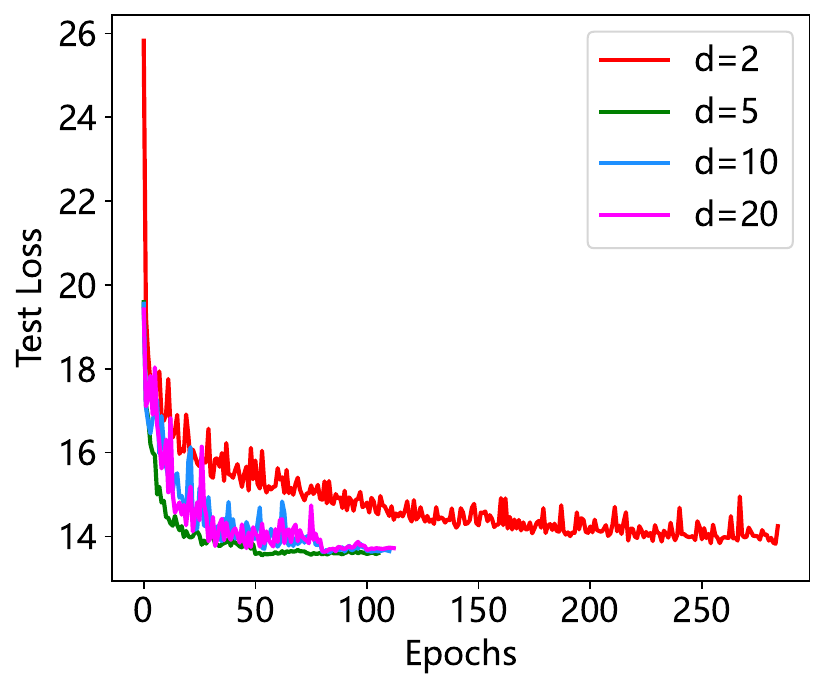}}
	\caption{Sensitivity analysis of parameter $d$ on PEMSD8.}
   \label{Sensitivity analysis of parameter $d$ on PEMSD8.}
\end{figure}

\subsection{Sensitivity to $d$}

The parameter \( d \) is set to \{2, 5, 10, 20\} to evaluate its sensitivity to model performance. The performance of the model with different values of \( d \) is shown in Figure \ref{Sensitivity analysis of parameter $d$ on PEMSD8.}. It is observed that \( d \) does not significantly affect model performance; however, it greatly impacts the training speed. When \( d \) is set between 5 and 10, the model trains quickly while maintaining performance. Therefore, \( d \) is recommended to be set around 5 to 10.

 \begin{figure}[ht]
	\centering  
	\subfigbottomskip=2pt 
	\subfigcapskip=-5pt 
	\subfigure[{Sensitivity of Parameter $M$ to PM-DMNet(P)}]{
        \label{Sensitivity of Parameter $d$ to PM-DMNet(P)}
		\includegraphics[width=0.48\linewidth]{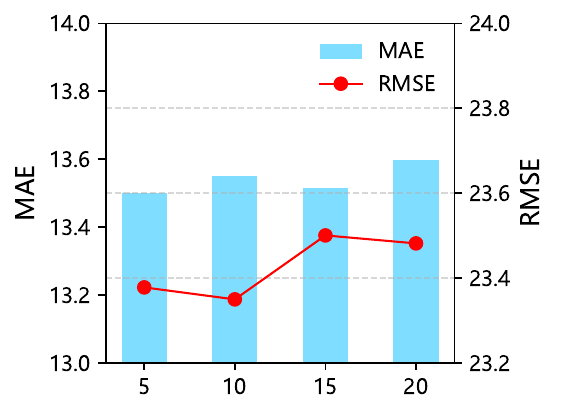}}
	\subfigure[{Sensitivity of Parameter $M$ to PM-DMNet(R)}]{
        \label{Sensitivity of Parameter $d$ to PM-DMNet(R)}
		\includegraphics[width=0.48\linewidth]{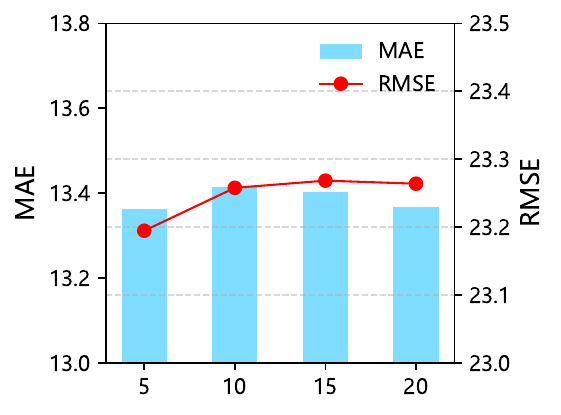}}
	\caption{Sensitivity analysis of parameter $M$ on PEMSD8.}
   \label{Sensitivity analysis of parameter $M$ on PEMSD8.}
\end{figure}

\subsection{Sensitivity to $M$}

The parameter \( M \) is set to \{5, 10, 15, 20\} to evaluate its sensitivity to model performance. The performance of the model with different values of \( M \) is shown in Figure \ref{Sensitivity analysis of parameter $M$ on PEMSD8.}. It is observed that the model achieves stable and excellent performance when \( M \) is between 5 and 20. Therefore, \( M \) is set to 10.

\section{Visualization}

To explore whether the Node Adaptive Parameter module captures the unique traffic patterns of each node, we utilize T-SNE \cite{van2008visualizing} to visualize the node embedding matrix $E$ used in the module trained on NYC-Taxi16 dataset.

\begin{figure}[ht]
  \centering
  \includegraphics[width=0.8\linewidth]{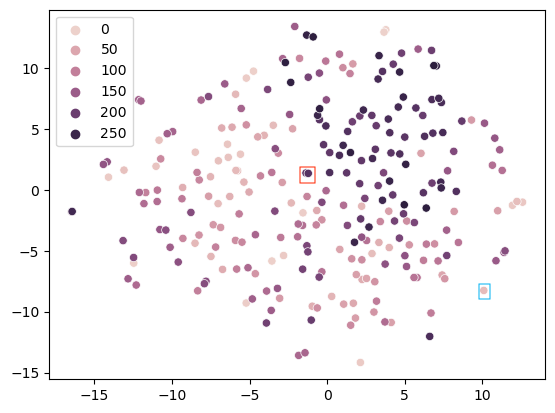}
  \caption{Visualization of node embeddings $E$. The nodes in the red-bordered area are nodes 215 and 222, the node in the blue-bordered area is node 26.}
  \label{Visualization of node embeddings $E$.}
\end{figure}

Figure \ref{Visualization of node embeddings $E$.} illustrates the visualization results of the node embeddings $E$. From the figure, it can be observed that certain nodes exhibit a clustering phenomenon, while a few nodes overlap, indicating high similarity in traffic patterns among them. Moreover, there are nodes that are far apart, suggesting significant differences in their traffic patterns. 

 \begin{figure}[ht]
	\centering  
	\subfigbottomskip=2pt 
	\subfigcapskip=-5pt 
	\subfigure[{Visualization of traffic demand for 'Pick-up' features}]{
        \label{Visualization of traffic demand for 'Pick-up' features}
		\includegraphics[width=\linewidth]{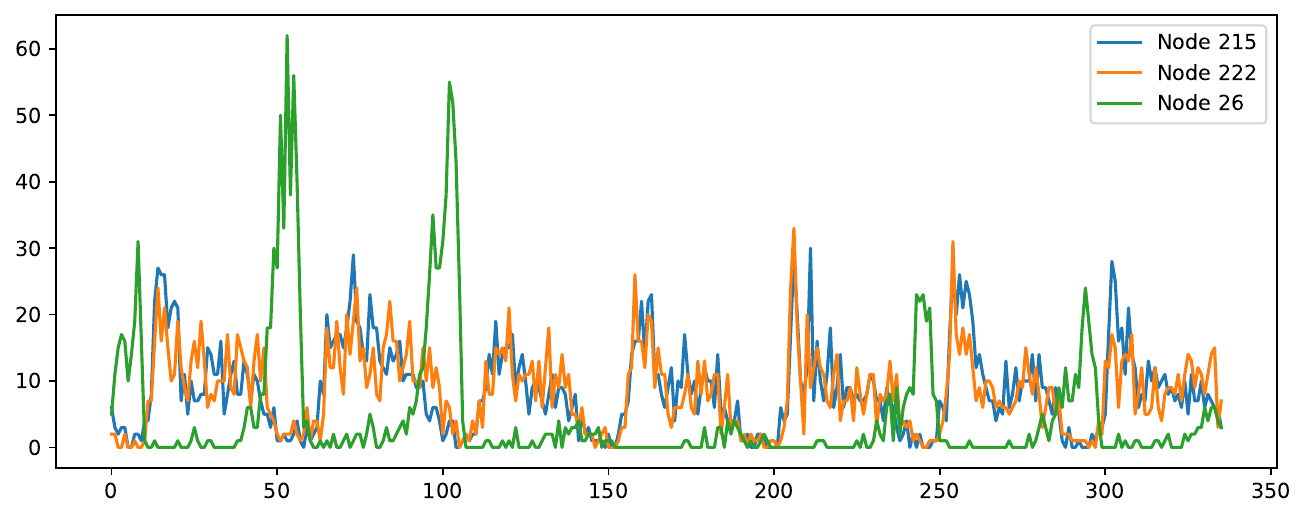}}
	\subfigure[{Visualization of traffic demand for 'Drop-off' features}]{
        \label{Visualization of traffic demand for 'Drop-off' features}
		\includegraphics[width=\linewidth]{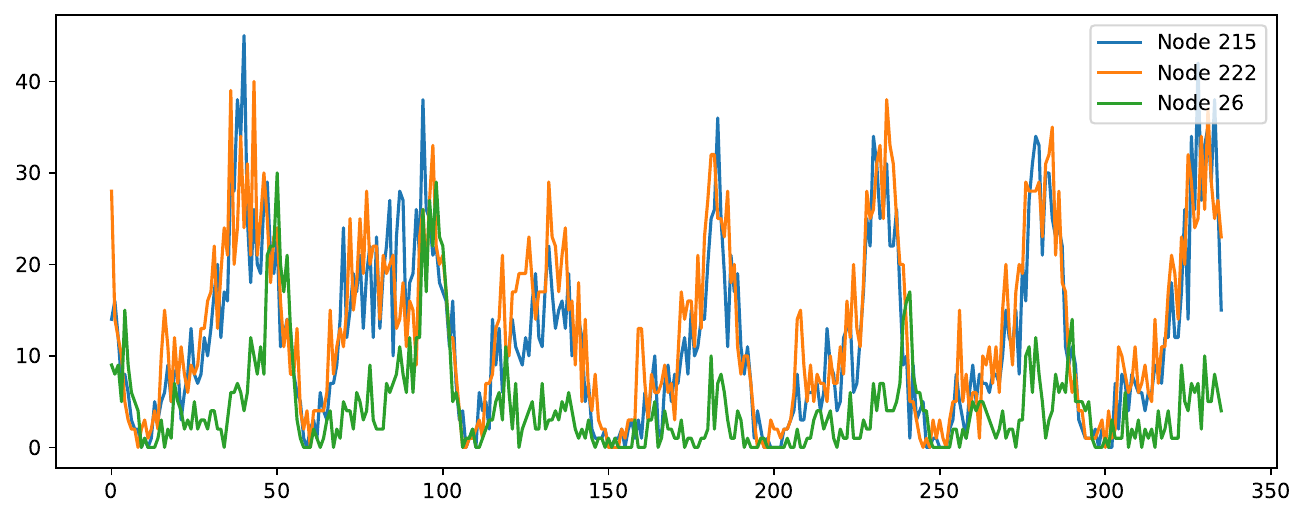}}
	\caption{Visualization of real traffic demand on NYC-Taxi16.}
   \label{Visualization of real traffic demand.}
\end{figure}

To further verify the high similarity in traffic patterns among nearby nodes and the differences in traffic patterns among distant nodes, we select adjacent nodes within the red-bordered area, specifically Node 215 and Node 222, as well as a distant node within the blue-bordered area, Node 26, for visualization of their traffic demand data.  Figures \ref{Visualization of traffic demand for 'Pick-up' features} and \ref{Visualization of traffic demand for 'Drop-off' features} respectively illustrate the trend changes in the 'Pick-up' and 'Drop-off' features of the traffic demand for these three nodes.  It is evident that the trends for Node 215 and Node 222 are highly similar, indicating a strong correlation between them.  Meanwhile, the trend for Node 26 is notably different from the other two nodes, suggesting a significant difference in their traffic patterns.  The visualization results above confirm that the Node Adaptive Parameter module can learn the traffic patterns of individual nodes effectively.

\section{CONCLUSION}

This paper proposes a novel traffic prediction model, PM-DMNet. PM-DMNet employs a new dynamic memory network module that learns the most representative traffic patterns into a memory network matrix. During prediction, the model extracts pattern features by matching the current traffic pattern with the memory network matrix. Additionally, PM-DMNet supports both parallel and sequential Multi-step prediction methods to meet different needs. To further enhance the accuracy of parallel Multi-step prediction, a transfer attention mechanism is introduced to mitigate the disparity between historical data and prediction targets. Extensive experiments validate the effectiveness of PM-DMNet. In future work, further methods for extracting features from patterns are planned to be explored.

\bibliographystyle{IEEEtran}
\bibliography{sample-base}

\begin{IEEEbiography}[{\includegraphics[width=1in,height=1.25in,clip,keepaspectratio]{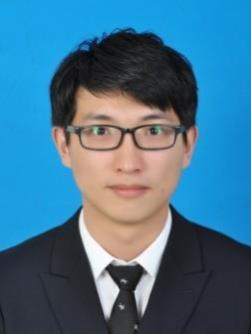}}]{Wenchao Weng}{\space}received his Bachelor’s degree in Information and Computing Science from Zhejiang Wanli University in 2019 and his Master’s degree in Computer Technology from Hangzhou Dianzi University in 2024. He is currently pursuing a Ph.D. in Computer Science and Technology at Zhejiang University of Technology. His research interests include data mining, spatio-temporal graph neural networks, and traffic forecasting.
\vspace{-10mm}
\end{IEEEbiography}

\begin{IEEEbiography}[{\includegraphics[width=1in,height=1.25in,clip,keepaspectratio]{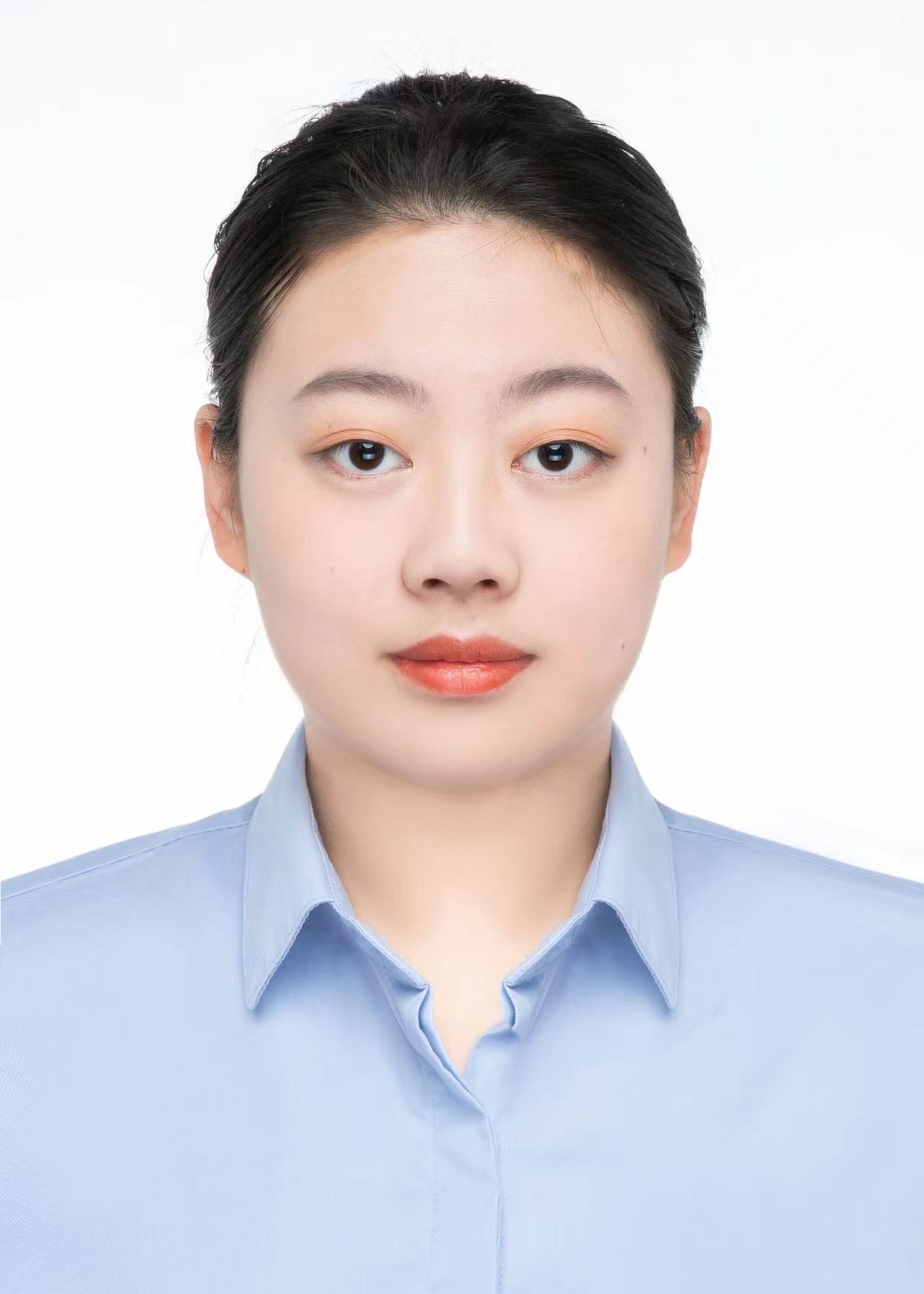}}]{Mei Wu}{\space}received the Bachelor's degree from Shandong University in China in 2022 and is currently pursuing a Master's degree in Computer Science at Hangzhou Dianzi University. Her main research interests focus on spatiotemporal graph data mining and intelligent transportation systems.
\vspace{-10mm}
\end{IEEEbiography}

\begin{IEEEbiography}[{\includegraphics[width=1in,height=1.25in,clip,keepaspectratio]{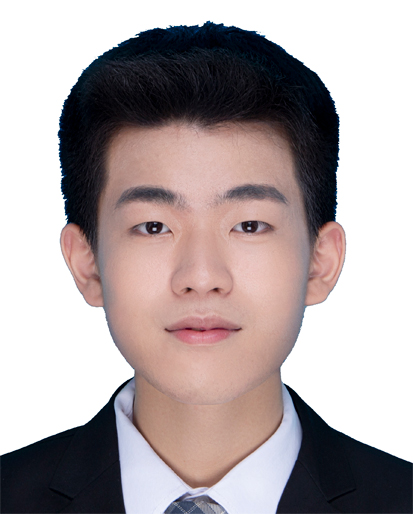}}]{Hanyu Jiang}{\space}is currently pursuing a Bachelor's degree at Hangzhou Dianzi University. His primary research focuses on the combination of bioinformatics and deep learning, specifically in the areas of multimodal and deep generative models.
\vspace{-10mm}
\end{IEEEbiography}

\begin{IEEEbiography}[{\includegraphics[width=1in,height=1.25in,clip,keepaspectratio]{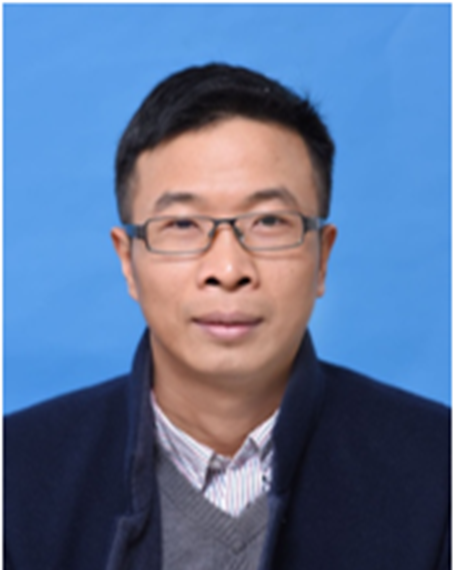}}]{Wanzeng Kong}{\space}(Senior Member, IEEE) received the Ph.D. degree from the Department of Electrical Engineering, Zhejiang University, in 2008. He was a Visiting Research Associate with the Department of Biomedical Engineering, University of Minnesota Twin Cities, Minneapolis, MN, USA, from 2012 to 2013. He is currently a Full Professor and the Director of the Cognitive Computing and BCI Laboratory, School of Computer Science and Technology, Hangzhou Dianzi University. His current research interests include machine learning, pattern recognition, and cognitive computing.

\vspace{-10mm}
\end{IEEEbiography}

\begin{IEEEbiography}[{\includegraphics[width=1in,height=1.25in,clip,keepaspectratio]{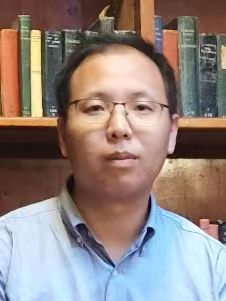}}]{Xiangjie Kong}{\space}(Senior Member, IEEE) received the B.Sc. and Ph.D. degrees from Zhejiang University, Hangzhou, China, in 2004 and 2009, respectively. He is a Professor with College of Computer Science and Technology, Zhejiang University of Technology, China. Previously, he was an Associate Professor with the School of Software, Dalian University of Technology, China. He has published over 200 scientific papers in international journals and conferences (with over 180 indexed by ISI SCIE). His research interests include urban computing, mobile computing, and computational social science. He is a Senior Member of the IEEE, a Distinguished Member of CCF, and is a member of ACM.
\vspace{-10mm}
\end{IEEEbiography}

\begin{IEEEbiography}[{\includegraphics[width=1in,height=1.25in,clip,keepaspectratio]{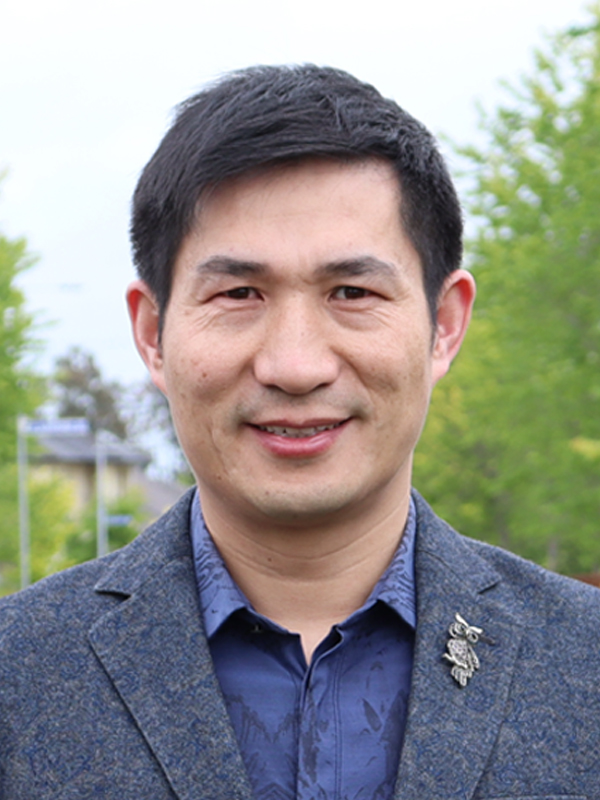}}]{Feng Xia}{\space}(Senior Member, IEEE) received the BSc and PhD degrees from Zhejiang University, Hangzhou, China. He is a Professor in School of Computing Technologies, RMIT University, Australia. Dr. Xia has published over 300 scientific papers in journals and conferences (such as IEEE TAI, TKDE, TNNLS, TC, TMC, TBD, TCSS, TNSE, TETCI, TETC, THMS, TVT, TITS, TASE, ACM TKDD, TIST, TWEB, TOMM, WWW, AAAI, ICLR, SIGIR, WSDM, CIKM, JCDL, EMNLP, and INFOCOM). His research interests include artificial intelligence, graph learning, brain science, digital health, and robotics. He is a Senior Member of IEEE and ACM, and an ACM Distinguished Speaker.

\end{IEEEbiography}

\end{document}